\documentclass[journal]{IEEEtran}
\renewcommand{\thesection}{\Roman{section}}

\renewcommand{\thesubsection}{\thesection.\Alph{subsection}}

\renewcommand{\thesubsubsection}{\thesubsection.\arabic{subsubsection}}

\usepackage{cite}

\ifCLASSINFOpdf
   \usepackage[pdftex]{graphicx}
\else
   \usepackage[dvips]{graphicx}
\fi
\usepackage{subcaption}
\usepackage{adjustbox}
\usepackage{wrapfig}

\usepackage[cmex10]{amsmath}
\usepackage{amsmath, amssymb, bm, xspace}
\usepackage[dvipsnames]{xcolor}
\usepackage{soul}
\sethlcolor{yellow}

\usepackage{array} 
\usepackage{float}
\newcolumntype{M}[1]{>{\centering\arraybackslash}m{#1}}

\usepackage{tabularx}
\usepackage{multirow}      
\usepackage{siunitx}       
\usepackage{csquotes}
\usepackage{url}

\usepackage[acronym]{glossaries}
\usepackage[colorinlistoftodos,prependcaption,textsize=tiny]{todonotes}

\usepackage{bbold}

\usepackage{booktabs}
\usepackage{lscape}

\usepackage{hyperref}
\hypersetup{colorlinks=true, linkcolor=black, citecolor=black, urlcolor=black}

\makeglossaries

\newacronym{dl}{DL}{deep learning}
\newacronym{ml}{ML}{machine learning}
\newacronym{ai}{AI}{artificial intelligence}
\newacronym{rs}{RS}{remote sensing}
\newacronym{eo}{EO}{earth observation}
\newacronym{cv}{CV}{computer vision}
\newacronym{xai}{xAI}{explainable artificial intelligence}
\newacronym{nn}{NN}{neural Networks}
\newacronym{bp}{BP}{backpropagation}
\newacronym{slc}{SLC}{single-label classification}
\newacronym{mlc}{MLC}{multi-label classification}
\newacronym{cnn}{CNN}{convolutional neural network}

\newacronym{lrp}{LRP}{Layer Relevance Propagation}
\newacronym{crp}{CRP}{Concept Relevance Propagation}
\newacronym{dtd}{DTD}{Deep Taylor Decomposition}
\newacronym{cam}{CAM}{Class Activation Mapping}
\newacronym{ggradcam}{Guided GradCAM}{Guided Gradient-weighted Class Activation Mapping}
\newacronym{hirescam}{HiResCAM}{High-Resolution Class Activation Mapping}
\newacronym{scorecam}{ScoreCAM}{Score-weighted Class Activation Mapping}
\newacronym{gradcam}{GradCAM}{Gradient-weighted Class Activation Mapping}
\newacronym{smcam}{SMCAM}{Self-Matching CAM}
\newacronym{lime}{LIME}{Local Interpretable Model-agnostic Explanations}
\newacronym{shap}{SHAP}{SHapley Additive exPlanations}
\newacronym{deeplift}{DeepLIFT}{Deep Learning Important FeaTures}
\newacronym{ig}{IG}{Integrated Gradients}
\newacronym{gbp}{GBP}{Guided Backpropagation}
\newacronym{rise}{RISE}{Randomised Input Sampling to provide Explanations}
\newacronym{occlusion}{Occlusion}{Occlusion}

\newacronym{map}{mAP}{mean Average Precision}
\newacronym{irof}{IROF}{Iterative Removal Of Features}
\newacronym{auc}{AUC}{Area Under the Curve}
\newacronym{aoc}{AOC}{Area Over the Curve}
\newacronym{ssim}{SSIM}{Structural Similarity Index Measure}
\newacronym{fc}{FC}{Faithfulness Correlation}
\newacronym{fe}{FE}{Faithfulness Estimate}
\newacronym{pf}{PF}{Pixel-Flipping}
\newacronym{rp}{RP}{Region Perturbation}
\newacronym{sel}{SEL}{Selectivity}
\newacronym{road}{ROAD}{RemOve And Debias}
\newacronym{sensN}{SENS-N}{Sensitivity-n}
\newacronym{as}{AS}{Average Sensitivity}
\newacronym{ms}{MS}{Maximum Sensitivity}
\newacronym{lle}{LLE}{Local Lipschitz Estimate}
\newacronym{ris}{RIS}{Relative Input Stability}
\newacronym{rrs}{RRS}{Relative Representation Stability}
\newacronym{ros}{ROS}{Relative Output Stability}
\newacronym{pg}{PG}{Pointing-Game}
\newacronym{tki}{TKI}{Top-K Intersection}
\newacronym{rma}{RMA}{Relevance Mass Accuracy}
\newacronym{rra}{RRA}{Relevance Rank Accuracy}
\newacronym{al}{AL}{Attribution Localisation}
\newacronym{mprt}{MPRT}{Efficient Model Parameter Randomization Test}
\newacronym{rl}{RL}{Random Logit}
\newacronym{sp}{SP}{Sparseness}
\newacronym{co}{CO}{Complexity}
\newacronym{eco}{ECO}{Effective Complexity}

\newacronym{ct}{Caltech101}{Caltech101}
\newacronym{dg}{DG-ML}{DeepGlobe Multi-Label}
\newacronym{ben}{BEN-v2.0}{BigEarthNet-S2 v2.0}
\newacronym{fbp}{FBP-ML}{Five Billion Pixel Multi-Label}

\newacronym{morf}{MoRF}{most relevant first}
\newacronym{lerf}{LeRF}{least relevant first}
\newacronym{mse}{RRR MSE}{Right for Right Reasons with Mean Squared Error}
\newacronym{ood}{OoD}{Out-of-Distribution}
\newacronym{slic}{SLIC}{Simple Linear Iterative Clustering}
\newacronym{ooi}{OoI}{Object of Interest}
\newacronym{sar}{SAR}{Synthetic Aperture Radar}
\newacronym{bpbased}{BP-based}{Backpropagation-based}

\newacronym{tsg}{TSG}{Texture Sensitivity Gap} 

\newacronym{nr}{NR}{Noise Resilience}
\newacronym{ar}{AR}{Adversary Reactivity}
\newacronym{iac}{IAC}{Intra-Consistency}
\newacronym{iec}{IEC}{Inter-Consistency}
\newacronym{mc}{MC}{Meta-Consistency}

\begin{document}

\newcommand{\predictor}{\ensuremath{\boldsymbol{f}}\xspace}

\newcommand{\explanator}{\ensuremath{\Phi}\xspace}
\newcommand{\metric}{\ensuremath{\Psi}\xspace}

\newcommand{\ex}{\ensuremath{\boldsymbol{e}}\xspace} 

\newcommand{\x}{\ensuremath{\boldsymbol{x}}\xspace}
\newcommand{\xbaseline}{\ensuremath{\overline{\boldsymbol{x}}}\xspace}
\newcommand{\xperturbed}{\ensuremath{\tilde{\boldsymbol{x}}}\xspace}

\newcommand{\xmasked}{\ensuremath{\boldsymbol{x}_{\left[\boldsymbol{x}_s=\overline{\boldsymbol{x}}_s\right]}}\xspace}
\newcommand{\xkmasked}{\ensuremath{\boldsymbol{x}_{\left[\boldsymbol{x}_k=\overline{\boldsymbol{x}}_k\right]}}\xspace}

\newcommand{\subsetOfIndices}{\ensuremath{s \in|S| \subseteq d}\xspace}
\newcommand{\kSubsetOfIndices}{\ensuremath{k \in|S| \subseteq d}\xspace}

\newcommand{\sumOverIndices}{\ensuremath{\sum_{i \in S}}\xspace}

\renewcommand{\as}[1]{\acrshort{#1}} 
\newcommand{\af}[1]{\acrfull{#1}}  
\newcommand{\al}[1]{\acrlong{#1}}  

%
\title{On the Effectiveness of Methods and Metrics for Explainable AI in Remote Sensing Image Scene Classification}
%
%
%

\author{Jonas~Klotz,~\IEEEmembership{Member,~IEEE,}
        Tom~Burgert,~\IEEEmembership{Member,~IEEE,}
        and~Begüm~Demir,~\IEEEmembership{Senior Member,~IEEE,}
        \thanks{Jonas Klotz, Tom~Burgert and Beg{\"u}m Demir are with the Faculty of Electrical Engineering and Computer Science, Technische Universit\"at Berlin, 10623 Berlin, Germany, also with the BIFOLD - Berlin Institute for the Foundations of Learning and Data, 10623 Berlin, Germany.
        Email: j.klotz@tu-berlin.de, t.burgert@tu-berlin.de, demir@tu-berlin.de.
        }

}

%
%

\markboth{Journal of \LaTeX\ Class Files,~Vol.~13, No.~9, September~2014}%
{Shell \MakeLowercase{\textit{et al.}}: Bare Demo of IEEEtran.cls for Journals}
%



\maketitle

\begin{abstract}

The development of explainable artificial intelligence (xAI) methods for scene classification problems has attracted great attention in remote sensing (RS). Most xAI methods and the related evaluation metrics in RS are initially developed for natural images considered in computer vision (CV), and their direct usage in RS may not be suitable. To address this issue, in this paper, we investigate the effectiveness of explanation methods and metrics in the context of RS image scene classification. 
In detail, we methodologically and experimentally analyze ten explanation metrics spanning five categories (faithfulness, robustness, localization, complexity, randomization), applied to five established feature attribution methods (Occlusion, LIME, GradCAM, LRP, and DeepLIFT) across three RS datasets.
Our methodological analysis identifies key limitations in both explanation methods and metrics. The performance of perturbation-based methods, such as Occlusion and LIME, heavily depends on perturbation baselines and spatial characteristics of RS scenes. Gradient-based approaches like GradCAM struggle when multiple labels are present in the same image, while some relevance propagation methods (LRP) can distribute relevance disproportionately relative to the spatial extent of classes. Analogously, we find limitations in evaluation metrics. Faithfulness metrics share the same problems as perturbation-based methods. Localization metrics and complexity metrics are unreliable for classes with a large spatial extent. In contrast, robustness metrics and randomization metrics consistently exhibit greater stability.
Our experimental results support these methodological findings.

Based on our analysis, we provide guidelines for selecting explanation methods, metrics, and hyperparameters in the context of RS image scene classification.
The code of this work will be publicly available at \url{https://git.tu-berlin.de/rsim/xai4rs}.

\end{abstract}


%
\IEEEpeerreviewmaketitle

\section{Introduction}
\label{sec:introduction}

\IEEEPARstart{T}{he} development of accurate methods for remote sensing (RS) image scene classification is one of the most important research topics in RS. In recent years, \af{dl} has emerged as the dominant approach for scene classification problems due to the high capability of DL models to extract and exploit the complex spatial and spectral content of RS images. RS image scenes can contain multiple land-use land-cover (LULC) classes and thus can be simultaneously associated with different class labels (i.e., multi-labels). Accordingly, the existing models are in general capable of assigning multiple LULC class labels to each RS image scene in an archive \cite{karalas2015deep, sumbul2020deep}. 
Despite their success, \as{dl} models often remain enigmatic black boxes, with almost no human understanding of their internal processes \cite{lipton2018mythos}. This lack of transparency raises concerns as decision-makers increasingly rely on these black-box systems \cite{saeed2023explainable}. New regulations for critical applications, such as the General Data Protection Regulation \cite{goodman2017european} and the Artificial Intelligence Act \cite{eu2023aiact}, require interpretability to ensure transparent decisions. Beyond regulatory concerns, effective interpretability can also improve model performance by addressing issues such as limited or biased training data, outliers, vulnerability to adversarial attacks, and overfitting \cite{saeed2023explainable}.

To this end, \af{xai} has emerged as a research area focused on revealing the factors behind the predictions of the \as{dl} model.  Although \as{xai} methods are widely used in the \as{cv} community \cite{schwalbe2024comprehensive}, they have only recently gained attention in \as{rs}. 
For a comprehensive overview of \as{xai} in \as{rs}, we refer the reader to \cite{hohl2024opening}. 
The most prominent type of explanation methods for \as{rs} scene classification problems are feature attribution methods \cite{hohl2024opening}. These methods assign importance scores to input features based on their contribution to the prediction of a model\cite{zeiler_visualizing_2013, ribeiro_why_2016, bach2015pixel, shrikumar_learning_2017, selvaraju2017grad}. While some feature attribution methods have been adapted to \as{rs} scene classification\cite{feng2021self, de2022towards}, most studies use feature attribution methods that were originally designed for image classification tasks on natural images \cite{yessou_comparative_2020, su2022cam, kucklick2023tackling}.

This raises the question of whether they are applicable to \as{rs} image scenes, as they exhibit different characteristics compared to natural images and can even be considered a distinct modality \cite{rolf2024mission}. 
To answer this question, it is essential to assess the quality of the generated explanations. Existing evaluation methods can be categorized into two groups: 1) qualitative assessments, which include domain expert evaluations or user studies; and 2) quantitative assessments, which employ explanation metrics \cite{doshi-velez_towards_2017}.  
Recently, Höhl et al. \cite{hohl2024opening} reviewed approximately 200 \as{rs} publications, finding that the vast majority evaluated qualitative, while only 16 studies included quantitative evaluation using explanation metrics.
This reliance on qualitative assessment raises concerns, as studies in the CV community have shown that qualitative assessment is insufficient for the robust verification of explanation methods \cite{murdoch2019definitions, nauta2023anecdotal}.

One of the key reasons for the limited use of quantitative evaluations in \as{rs} is the absence of clear guidelines to assess the quality of the explanation in this domain. The need for such guidelines is underscored by the inherent difficulty of quantitative evaluation, since the explanation metrics depend heavily on their parameterization \cite{hooker2019benchmark, nauta2023anecdotal}. Moreover, the lack of ground truth for the explanations renders them unverifiable \cite{hedstrom2023meta}, requiring the use of a wide range of metrics for a more comprehensive evaluation \cite{hedstrom2023quantus}. In many \as{rs} studies, only a small part of this range is analyzed, often focusing on single aspects, such as localization capability \cite{su2022cam}, while only a few studies have evaluated a broader variety of metrics \cite{kakogeorgiou_evaluating_2021, mohan2023quantitative}. 
In particular, none of these studies have addressed the fact that most explanation metrics have been originally designed for natural images and can be limited when applied to \as{rs} scene classification.

To address these issues, we present the first unified evaluation of the explanation methods and metrics for \as{rs} image scene classification. We assess five widely used feature attribution methods: \as{occlusion} \cite{zeiler_visualizing_2013}, \as{lime} \cite{ribeiro_why_2016}, \as{lrp} \cite{bach2015pixel}, \as{deeplift} \cite{shrikumar_learning_2017} and \as{gradcam} \cite{selvaraju2017grad};
and evaluate ten explanation metrics spanning five categories: faithfulness (\as{fe} \cite{alvarez_melis_towards_2018}, \as{irof} \cite{rieger_irof_2020}), robustness (\as{as} \cite{yeh_fidelity_2019}, \as{lle} \cite{alvarez_melis_towards_2018}), localization (\as{rra} \cite{arras2022clevr}, \as{tki} \cite{theiner_interpretable_2022}), complexity (\as{sp} \cite{chalasani_concise_2020}, \as{co} \cite{bhatt_evaluating_2020}) and randomization-based (\as{mprt} \cite{adebayo_sanity_2018}, \as{rl} \cite{sixt_when_2020}). These methods and metrics are selected for their popularity in \as{rs} and to cover a broad spectrum of evaluation criteria.
We perform a methodological analysis to examine whether the existing explanation methods and metrics are suitable for the classification of \as{rs} images. 
Then, we perform an extensive experimental evaluation using MetaQuantus \cite{hedstrom2023meta} to assess the reliability and effectiveness of the explanation metrics under minor perturbations and disruptive changes, supporting our methodological findings with experimental validation.

We would like to note that, unlike previous work that primarily surveys the state of the art in \as{xai} for \as{rs} \cite{hohl2024opening}, our study goes beyond a methodological overview by conducting a systematic, in-depth analysis of explanation methods and evaluation metrics. In particular, we critically examine the reliability of explanation metrics, which is an aspect that, to the best of our knowledge, has not previously been investigated in the \as{rs} domain.
The primary contributions of this work are summarized as follows:
\begin{itemize}

\item We provide an extensive methodological analysis of how the differences between \as{rs} and natural images impact the effectiveness of existing feature attribution methods and evaluation metrics in the context of scene classification in RS. 
\item Based on this analysis, we conduct an experimental study of the selected attribution metrics and methods across multiple \as{rs} scene classification datasets.
\item Based on our methodological and experimental findings, we develop practical guidelines to assist \as{rs} practitioners in selecting appropriate explanation methods and evaluation metrics for \as{rs} image scene classification and recommend suitable explanation methods and metrics. Our pipeline can be reproduced and adapted for different datasets or classification tasks in \as{rs}, providing a valuable resource for the RS community.
\end{itemize}

The remainder of this paper is organized as follows.
Section~\ref{sec:theory_methods} and Section~\ref{sec:theory_metrics} provide a methodological evaluation of the effectiveness of explanation methods and explanation metrics for \as{rs} image scene classification, respectively.
In Section~\ref{sec:data_experiments} we describe the datasets and the experimental setup. The results of our experiments are presented in Section~\ref{sec:results}. 
Finally, Section~\ref{sec:conclusion_discussion} draws the conclusion of the work.

\begin{figure*}[t]
    \centering
    \includegraphics[width=0.7\linewidth]{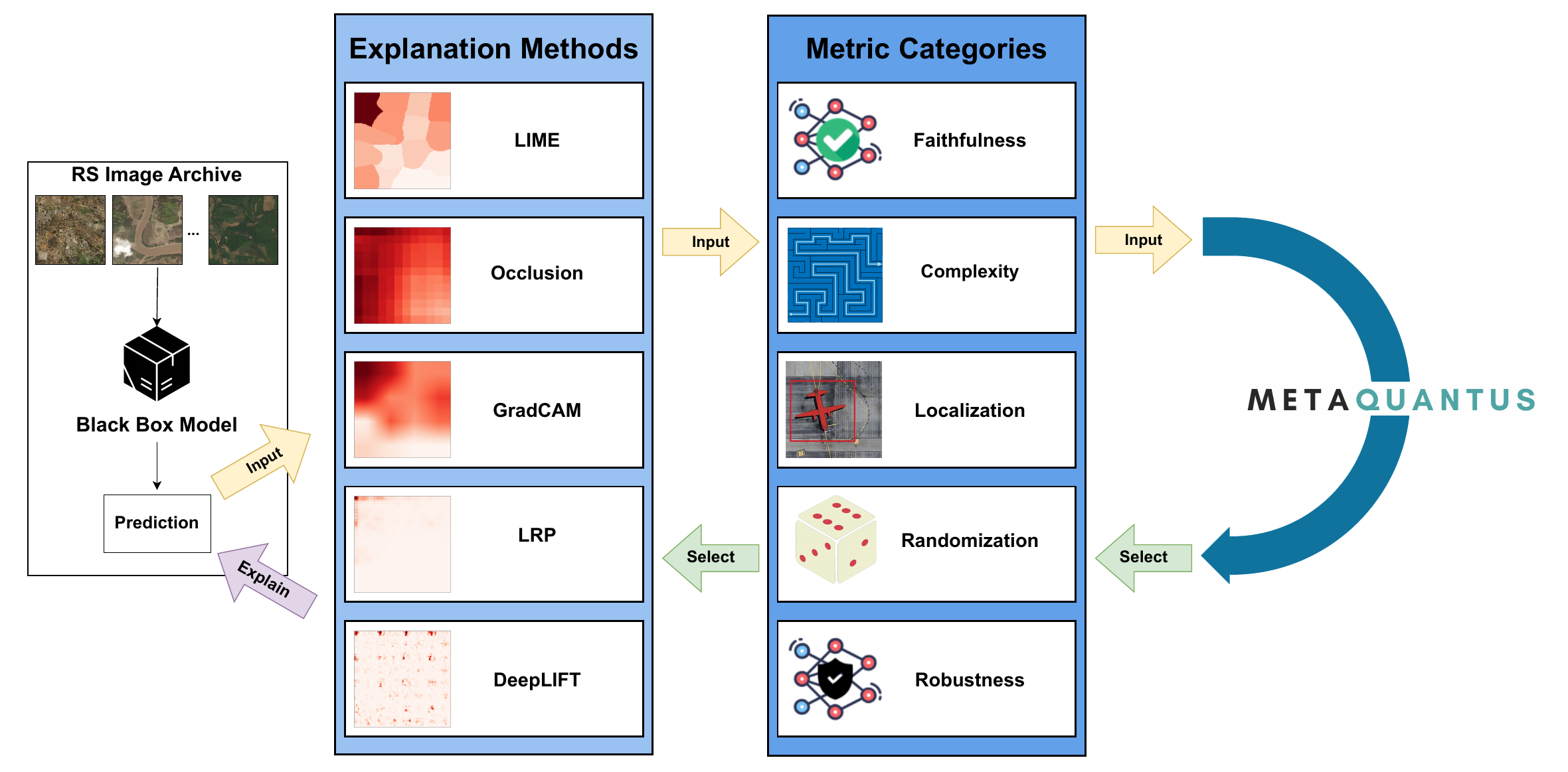}
    \caption{The evaluation protocol for selecting appropriate explanation methods and metrics for \as{rs} image scene classification. It involves generating explanations using various feature attribution methods, estimating their quality using selected metrics, assessing the reliability of the metric with MetaQuantus \cite{hedstrom2023meta}, and using the most reliable metrics to select the most suitable explanation methods.}
    \label{fig:main-figure}
\end{figure*}

\section{Explanation methods for RS image scene classification}
\label{sec:theory_methods}
\noindent
In this section, we provide a methodological analysis of the considered explanation methods with respect to the specific challenges of \as{rs} images. Most \as{xai} research has focused on natural images, which differ significantly from \as{rs} images \cite{rolf2024mission}. In this paper, we examine commonly applied explanation methods in \as{rs} scene classification, focusing on methods that generate feature attribution maps, since these methods are most widely used in \as{rs} \cite{hohl2024opening}. However, we note that alternative paradigms such as Multiple Instance Learning \mbox{\cite{li2020deep, bi2019multiple, bi2022all}} or counterfactual explanations \mbox{\cite{dantas2023towards}} have also been utilized in RS.
Feature attribution methods can be grouped into perturbation-based and backpropagation-based methods \cite{bai_explainable_2021}. Perturbation methods modify inputs to measure output changes, while backpropagation methods propagate values from outputs to inputs, typically gradients or relevancy scores. We consider three categories: i) \textit{perturbation-based}, ii) \textit{relevancy-propagation-based}, and iii) \textit{gradient-propagation-based} methods.

Our analysis covers five representative methods: 
1) two perturbation-based methods (\as{occlusion} \cite{zeiler_visualizing_2013}, \as{lime} \cite{ribeiro_why_2016});
2) two relevancy-propagation-based methods (\as{lrp} \cite{bach2015pixel}, \as{deeplift}; \cite{shrikumar_learning_2017}) and 
3) one gradient-propagation-based method (\as{gradcam} \cite{selvaraju2017grad}).
For each category, we first provide a general description, followed by a definition of the explanation methods and a discussion of the limitations considering the differences between \as{rs} and natural images. 

Let us consider a \as{rs} scene classification problem, where a DL model $f$ is trained on an archive $\boldsymbol{X}_{\mathrm{tr}}$ of $N$ pairs $(\boldsymbol{x}_i,\boldsymbol{y}_i)_{i=1}^N$ for multi-label classification (the single-label case is omitted for simplicity). Each image $\boldsymbol{x} \in \mathbb{R}^{H \times W \times C}$ represents a scene captured by an airborne or satellite sensor, with $C$ spectral bands and $H,W$ spatial dimensions. The model $f$ maps an input image to predicted labels $  f(\boldsymbol{x} ) = \hat{\boldsymbol{y}}$,
where $ \hat{\boldsymbol{y}} $ represents the predicted class probabilities or binary assignments. 
To interpret the reasoning behind the prediction of a model $f$ for an input $\boldsymbol{x}$, post-hoc explanation methods, such as feature attribution methods $\Phi$, can be applied:
\begin{equation}
    \Phi(\boldsymbol{x}, f, \hat{\boldsymbol{y}} ) = \hat{\boldsymbol{e}},
\end{equation}
where $ \Phi $ distributes relevancy scores $r$ across $ \boldsymbol{x} $, resulting in an attribution map $ \hat{\boldsymbol{e}} \in \mathbb{R}^{ H \times W \times C}$. In practice, $ \hat{\boldsymbol{e}}$ is often aggregated across spectral bands.
For simplicity, we omit explicit parameterization unless necessary.
\subsection{Perturbation-based methods}
\subsubsection{Background}
Perturbation-based methods explain predictions by measuring output changes between original and perturbed inputs.  
Let $\mathbb{X} \subseteq \mathbb{R}^{H \times W \times C}$ denote the input space and $\Theta$ a set of perturbation parameters.  
A perturbation function $\mathcal{P}_{\mathbb{X}} : \mathbb{X} \times \Theta \;\to\; \mathbb{X}$ is defined as:
\begin{equation}
\mathcal{P}_{\mathbb{X}}(\boldsymbol{x}, \theta)  \;=\;  \tilde{\boldsymbol{x}}
\end{equation}
with $\boldsymbol{x} \in \mathbb{X}$ and $\theta \in \Theta$. In other words, $\mathcal{P}_{\mathbb{X}}(\boldsymbol{x}, \theta)$ modifies $\boldsymbol{x}$ according to the perturbation parameter $\theta$, while still preserving the overall structure such that $\tilde{\boldsymbol{x}} $ is a valid point in the same space.
%
Common perturbation functions are blurring \cite{kang2019interpreting}, introducing noise \cite{ yeh_fidelity_2019}, or replacing parts of the image \cite{agarwal_rethinking_2022}. 
We consider \as{occlusion} \cite{zeiler_visualizing_2013} and \as{lime} \cite{ribeiro_why_2016} as perturbation-based methods.

In the \as{occlusion} \cite{zeiler_visualizing_2013} method, each $\theta \in \Theta$ specifies a region (mask) $\Omega \subseteq \{1,\dots,H\}\times\{1,\dots,W\}$ and a baseline $\alpha \in \mathbb{R}$. The perturbation function is:
        \begin{equation}
          \bigl[\mathcal{P}_{\mathbb{X}}(\boldsymbol{x},\theta)\bigr]_{j,k}
          \;=\;
          \begin{cases}
             \alpha, & \text{if }(j,k)\in \Omega,\\
             \boldsymbol{x}_{j,k}, & \text{otherwise}.
          \end{cases}
        \end{equation}
Then, we calculate the prediction certainty difference $\Delta s_c$ as:
\begin{equation}
              \Delta s_c(\theta) = f_c(\boldsymbol{x}) - f_c\bigl(\mathcal{P}_{\mathbb{X}}(\boldsymbol{x},\theta)\bigr),
\end{equation}
for each mask $\Omega$. A large positive $\Delta s_c(\theta)$ means that the perturbation significantly lowered the prediction certainty of the model for class $c$, suggesting that the perturbed features are important. Then, we aggregate the $\Delta s_c(\theta)$ values to produce an attribution map that assigns high importance to the regions whose occlusion most reduces the model prediction certainty for class $c$. 
Thus, the \as{occlusion} method $\Phi_{\text{Occlusion}}$ is defined as:
\begin{equation}
\label{eq:final_phi}
\Phi_{\text{Occlusion}}(j,k)
\;=\;
\frac{1}{\bigl|\{\theta : (j,k) \in \Omega\}\bigr|}
\sum_{\theta : (j,k)\in \Omega} \Delta s_c(\theta).
\end{equation}

The \as{lime} method $\Phi_{\mathrm{LIME}}$ explains a black-box model $f$ by locally approximating it with an interpretable model trained on binary-encoded perturbations of the input.  
Let $S = \{S_1, S_2,\dots,S_M\}$ be a segmentation of $\boldsymbol{x}$ into $M$ superpixels. We define $\boldsymbol{z} \in \{0,1\}^M$ as a binary vector indicating the presence (1) or absence (0) of superpixels. The perturbation function is:
\begin{equation}
\mathcal{P}_{\mathbb{X}} : \mathbb{X} \times \{0,1\}^M \;\to\; \mathbb{X},
\end{equation}
such that, for each $\theta = \boldsymbol{z}$, where:
\begin{equation}
 \mathcal{P}_{\mathbb{X}}(\boldsymbol{x}, \boldsymbol{z})_{j,k}
 \;=\;
 \begin{cases}
  [\boldsymbol{x}]_{j,k}, & \text{if } (j,k)\in S_m \text{ ,} \boldsymbol{z}_m=1,\\
  \alpha,                & \text{if } (j,k)\in S_m \text{ , } \boldsymbol{z}_m=0,
 \end{cases}
\end{equation}
where $\alpha \in \mathbb{R}^C$ is a constant baseline. Varying $\boldsymbol{z}_m$ generates perturbed samples around $\boldsymbol{x}$.  
Then, we fit a simple model $g$ that approximates $f$ near $\boldsymbol{x}$:
\begin{equation}
\Phi_{\mathrm{LIME}}(\boldsymbol{x}) 
\;=\; 
\operatorname{argmin}_{g \in \mathcal{G}} 
\Bigl[
 \mathcal{L}\bigl(f,g,\Pi_{x}\bigr)
 \;+\;
\Upsilon(g)
\Bigr],
\end{equation}
where $\mathcal{L}$ is a loss, $\Upsilon(g)$ a complexity regularizer (e.g., sparsity or depth), and $\Pi_x(\tilde{\boldsymbol{x}})$ a proximity measure quantifying distance of $\tilde{\boldsymbol{x}}$ to $\boldsymbol{x}$.

\subsubsection{Limitations}
\label{subsubsec:perturbation_method_problems}
\begin{figure}[ht]
    \centering
    \begin{subfigure}[t]{0.15\textwidth}
        \centering
        \includegraphics[width=\textwidth]{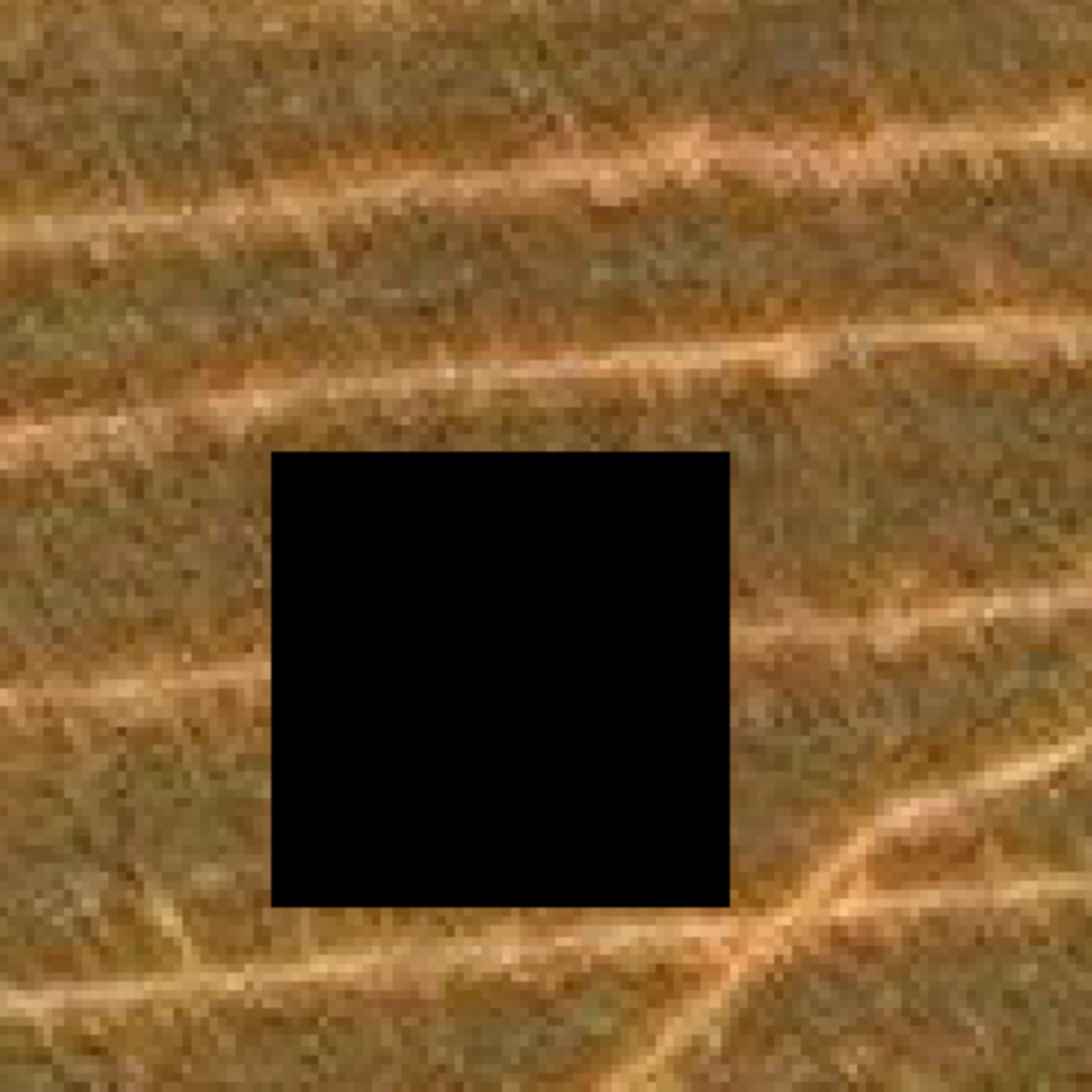}
        \caption{}
        \label{fig:methods_occlusion_a}
    \end{subfigure}
    \hspace{0.5cm}
    \begin{subfigure}[t]{0.15\textwidth}
        \centering
        \includegraphics[width=\textwidth]{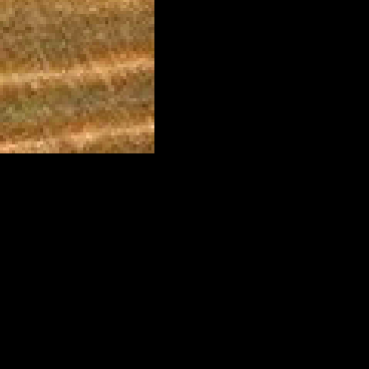}
        \caption{}
        \label{fig:methods_occlusion_b}
    \end{subfigure}
    \caption{Visualizations of perturbed images using $\alpha =0$ from the \as{dg} \cite{DeepGlobe18} dataset for the target class $c$ (\textit{agricultural land}) and its prediction certainty for the original image: $f_{c}(\boldsymbol{{x}}) = 0.9$.  
    a) Perturbed sample, where $f_{c}(\boldsymbol{\tilde{x}}) = 0.99$:
    b) Perturbed sample, where $f_{c}(\boldsymbol{\tilde{x}}) = 0.38$.
    }
    \label{fig:methods_occlusion}
\end{figure}

A key component of perturbation-based methods is the perturbation baseline, which represents the absence of the occluded input feature. In practice, many \as{rs} studies use a scalar value $\alpha$, often zero, as in Kakageorgiou et al. \cite{kakogeorgiou_evaluating_2021}. However, scalar baselines can produce \af{ood} samples \cite{qiu2022generating}, and \as{dl} models yield unreliable predictions on inputs outside their training distribution \cite{nguyen2015deep}.  
Fig.~\ref{fig:methods_occlusion_a} illustrates this limitation with an RGB image from the \af{dg} dataset \cite{DeepGlobe18}. A rectangular patch is perturbed with the \as{occlusion} method, using a black perturbation baseline ($\alpha=0$). Despite occluding a part of the class, the prediction of the model for the class \textit{agricultural land} increased from $f_c(\boldsymbol{x}) = 0.90$ on the original input to $f_c(\tilde{\boldsymbol{x}}) = 0.99$ after the occlusion. Even with most pixels perturbed (Fig.~\ref{fig:methods_occlusion_b}), the model outputs $f_c(\tilde{\boldsymbol{x}})=0.38$, highlighting the unpredictable behavior introduced by a poorly chosen perturbation baseline.

This challenge is amplified in multispectral \as{rs} images, where spectral bands capture distinct information. Common baselines, such as all-zero vectors, are unsuitable because some bands rarely take zero values in operational conditions. For example, in thermal images, a zero baseline has no physical meaning, and in near-infrared bands, some radiation is always present. Meaningful baselines are required to account for spectral characteristics.

Furthermore, the regions within each RS image are associated with different LULC classes, and unlike in natural images, background content does not exist. Another limitation arises because \as{rs} images contain multiple LULC classes without background content. Let $A_c(\boldsymbol{x})$ denote the pixel count of class $c$ in $\boldsymbol{x}$ and $A(\Omega)$ the area of an occlusion mask. If $A(\Omega) < A_c(\boldsymbol{x})$, the perturbation may not obscure enough pixels to lower prediction certainty. Fig.~\ref{fig:methods_occlusion_a} shows such a case, where occlusion fails to reduce $f_c(\tilde{\boldsymbol{x}})$. 
This limitation is less pronounced in natural images, where discriminative features are often unique (e.g., silhouette, object parts). In such cases, occluding typically leads to a noticeable reduction in prediction certainty even if $A_c(\boldsymbol{x}) > A(\Omega)$.

A similar problem occurs for the \as{lime} method, where $\mathcal{P}_{\mathbb{X}}$ partitions $\boldsymbol{x}$ in different superpixels and partially occludes them.  Unless most superpixels are replaced (Fig.~\ref{fig:methods_occlusion_b}), the prediction remains unchanged, resulting in a linear model trained primarily on positive class predictions. 
This imbalance causes the estimated weights of the linear model to be unreliable, leading to an inaccurate interpretation of feature importance. 
Since no single superpixel strongly influences the model prediction, \as{lime} assigns nearly equal importance across superpixels, underestimating discriminative features due to regularization. 
Additionally, the more pixels are replaced by the perturbation baseline, the higher the proximity measure $\Pi_x$ due to the large distance from the original image. As a result, individual pixels are incorrectly attributed as less relevant. 

\subsection{Gradient-propagation-based methods}
\subsubsection{Background}

Gradient-propagation-based methods use gradients to explain model predictions. We consider \as{gradcam} \cite{selvaraju2017grad}, which utilizes gradients flowing into a convolutional layer, typically the last one, due to its semantic and spatial content. For class $c$, the gradients of the class score $Y^c$ with respect to activations $A^p$ of layer $l$ yield importance weights $\omega^c_p$:
\begin{equation}
\omega^c_p = \frac{1}{Z} \sum_{j,k} \frac{\partial Y^c}{\partial A^p_{j,k}},
\end{equation}
where global averaging collapses spatial locations $(j,k)$. These weights approximate the contribution of each feature map $p$ to class $c$. The attribution map is the weighted feature maps followed by ReLU, ensuring only positive contributions are visualized:
\begin{equation}
    \Phi_{\mathrm{GradCAM}}^c = \text{ReLU}\left(\sum_p \omega^c_p A^p \right).
\end{equation}
\subsubsection{Limitations}
Gradient-propagation-based methods exhibit limitations when applied to \as{rs} scenes with multiple LULC classes. For instance, in a scene containing five airplanes, \as{gradcam} localizes only a single instance \cite{huang2021better}, underscoring its inability to resolve multiple objects of interest at varying scales \cite{fu2019multicam, wang2020weakly}. This failure originates from the reliance of \as{gradcam} on a global gradient averaging mechanism, which collapses spatial gradient information into a single channel-wise importance weight. For two distinct objects \( O_1 \) and \( O_2 \), activating disjoint regions \( R_1 \) and \( R_2 \) in \( A^p \), the gradients \(\nabla_1\) and \(\nabla_2\) are:  
\begin{equation}
    \frac{\partial Y^c}{\partial A_{j,k}^p} = 
\begin{cases} 
\nabla_1, & \forall (j,k) \in R_1 \\
\nabla_2, & \forall (j,k) \in R_2 \\
0, & \text{otherwise}.
\end{cases}
\end{equation}
Assuming equal spatial extents (\(\lvert R_1\rvert = \lvert R_2\rvert = N\)) and total positions \(Z = u \times v\), \as{gradcam} computes the channel weight as:  
\begin{equation}
\omega_p^c = \frac{N}{Z}\,\bigl(\nabla_1 + \nabla_2\bigr).
\end{equation}
When $\nabla_1 \neq \nabla_2$, the larger gradient contributes more to the sum, so the computed weight is more reflective of the larger gradient. Since $\omega_p^c$ is a scalar applied uniformly to $A^p$, spatially distinct contributions from $R_1$ and $R_2$ merge, and the attribution map $\Phi_{\mathrm{GradCAM}}^c$ loses the granularity needed to separate multiple objects.


\subsection{Relevancy-propagation-based methods}
\subsubsection{Background}
Relevancy-propagation-based methods explain predictions by back-propagating relevance scores to attribute importance to input features.  
We consider \as{deeplift} \cite{shrikumar_learning_2017} and \as{lrp} \cite{bach2015pixel}.

\as{deeplift} \cite{shrikumar_learning_2017} attributes the output difference $f_c(\boldsymbol{x})-f_c(\tilde{\boldsymbol{x}})$ to input differences between $\boldsymbol{x} \in \mathbb{R}^{H \times W \times C}$ and a baseline $\tilde{\boldsymbol{x}}=\mathcal{P}_{\mathbb{X}}(\boldsymbol{x},\theta)$. Forward passes are run for both inputs, and the weighted activation difference is stored as $\Delta z_{s,t}=w_{s,t}^{(l, l+1)}(\boldsymbol{x}_s^l-\tilde{\boldsymbol{x}}^{l}_s)$, where $\boldsymbol{x}_s^l$ and $\tilde{\boldsymbol{x}}^{l}_s$ are neuron activations in layer $l$, and $w_{s,t}^{(l,l+1)}$ is the weight to neuron $t$ in layer $l+1$. The relevance backpropagation rules are:
\begin{equation}
r_t^{(L)} =
\begin{cases}
  f_c(\boldsymbol{x}) - f_c(\tilde{\boldsymbol{x}}), & t = c,\\
  0, & t \neq c,
\end{cases}
\end{equation}
and
\begin{equation}
r_s^{(l)}
= \sum_{t}
  \frac{\Delta z_{s,t}}{\displaystyle\sum_{s'}\Delta z_{s',t}}
  \;r_t^{(l+1)},
\end{equation}
and the final attribution $\Phi_{\text{DeepLift}}$ is defined as:
\begin{equation}
\Phi_{\text{DeepLift}}(f_c,\boldsymbol{x})_s = r_s^{(1)},
\end{equation}
where $L$ is the output layer and $r_s^{(1)}$ is the contribution of each feature on the input layer. For further propagation rules, we refer the reader to \cite{shrikumar_learning_2017}.

\noindent\as{lrp} \cite{bach2015pixel} propagates output relevance back to inputs by interpreting the network as a flow graph. For a layer $l$, relevance scores $r$ are redistributed as follows:
\begin{equation}
r_s^{(l)} \;=\; \sum_{t}
  \frac{z_{s,t}}{\displaystyle\sum_{s'} z_{s',t}}
  \;r_t^{(l+1)}\,,
\end{equation}
where $z_{s,t}$ is the contribution of neuron $s$ in layer $l$ to neuron $t$ in $l+1$, ensuring conservation of total relevance. Specific propagation rules depend on the model architecture and task \cite{montavon_layer-wise_2019}. For the final layer, $r_t^{(L)}$ is initialized as the target class score $f_c(\boldsymbol{x})$. The input attribution \as{lrp} method $\Phi_{\text{LRP}}$ is then:
\begin{equation}
\Phi_{\text{LRP}}\bigl(f_c,\boldsymbol{x}\bigr)_s \;=\; r_s^{(1)}.
\end{equation}

\begin{figure}[!t]
    \centering
    \begin{subfigure}[!t]{0.15\textwidth}
        \includegraphics[width=\textwidth]{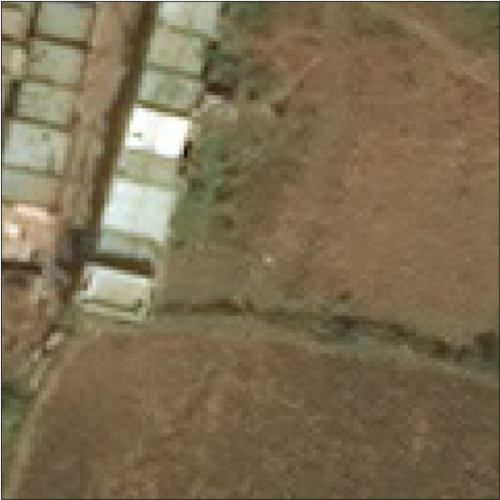}
        \caption{}
        \label{fig:lrp_original}
    \end{subfigure}
    \begin{subfigure}[!t]{0.15\textwidth}
        \includegraphics[width=\textwidth]{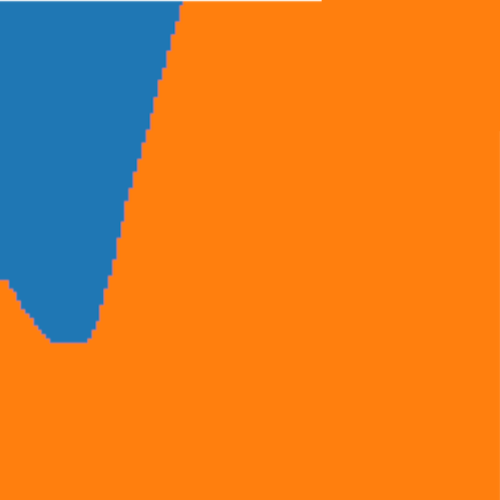}
        \caption{}
        \label{fig:lrp_reference}
    \end{subfigure}
    \\
    \begin{subfigure}[!t]{0.15\textwidth}
        \includegraphics[width=\textwidth]{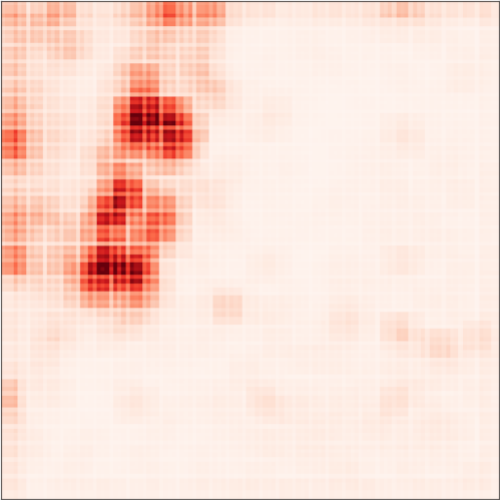}
        \caption{}
    \end{subfigure}
    \begin{subfigure}[!t]{0.15\textwidth}
        \includegraphics[width=\textwidth]{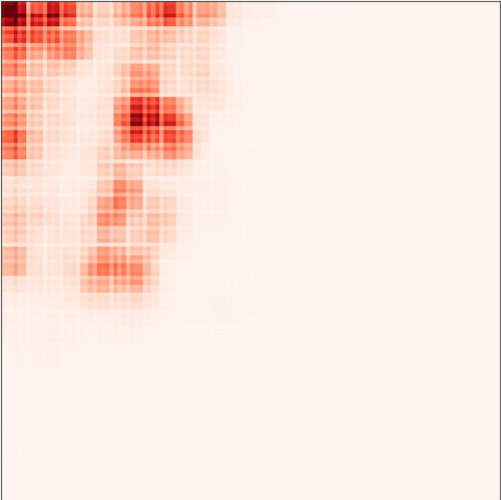}
        \caption{}
    \end{subfigure}
    \caption{LRP visualizations for an image from the \as{dg} \cite{DeepGlobe18} dataset.  
    a) Original image with two classes: Urban Land and Agricultural Land;  
    b) Pixel-wise reference map (blue: Urban Land, orange: Agricultural Land);  
    c) LRP explanation for Agricultural Land (normalized range: [0,1], unnormalized range: [0,0.02]);  
    d) LRP explanation for Urban Land (normalized range: [0,1], unnormalized range: [0,0.18]).}
    \label{fig:lrp_global_conservation}

\end{figure}

\subsubsection{Limitations}
Since \as{deeplift} computes relevance as $f_c(\boldsymbol{x})-f_c(\tilde{\boldsymbol{x}})$, it inherits the baseline-dependence limitations of perturbation-based methods (see Section~\ref{subsubsec:perturbation_method_problems}). Sturmfels et al.\ \cite{sturmfels2020visualizing} studied baseline choices for \acrlong{ig}, which also relies on a difference-from-reference formulation, testing Gaussian blur, uniform values, maximum distance, and black baselines. They show that suitable baselines require domain-specific knowledge to represent missingness correctly. This problem is intensified in \as{rs} scene classification, where multiple spectral bands complicate the selection of a meaningful baseline.

A central property of \as{lrp} and related methods such as \al{ig} \cite{sundararajan_axiomatic_2017} is \emph{relevance conservation}: the total input relevance equals the network output, $\sum_j r_j^{c}=f_c(\boldsymbol{x})$ \cite{montavon_layer-wise_2019}. 
\as{rs} scenes usually contain several LULC classes, so that a single class may occur in multiple, spatially disconnected regions. The discriminative information is therefore distributed rather than localized. 
Because the overall relevance budget is fixed, two classes that yield comparable output scores but occupy differently sized areas receive different relevance densities. The larger region is attributed a lower per-pixel relevance than the smaller region.
Fig.~\ref{fig:lrp_global_conservation} shows this discrepancy for a \as{dg} image containing \textit{agricultural land} ($c_{\text{major}}$) and \textit{urban land} ($c_{\text{minor}}$). Since $f_{c_{\text{major}}}(\boldsymbol{x}) \approxeq f_{c_{\text{minor}}}(\boldsymbol{x})$, both propagate similar total relevance. However, the larger $c_{\text{major}}$ region yields lower per-pixel relevance ($[0,0.02]$) than $c_{\text{minor}}$ ($[0,0.18]$). As attributions are typically normalized to $[0,1]$, this discrepancy increases sensitivity to outliers.

\section{Explanation metrics for RS image scene classification}
\label{sec:theory_metrics}
In this section, we first provide a brief overview of the categorization of explanation metrics. We then introduce selected explanation metrics from different categories and discuss their suitability to estimate the quality of explanations for \as{rs} image scene classification. Explanation metrics quantify the quality of explanation methods. An explanation is commonly defined as a \enquote{presentation [...] in human-understandable terms} \cite{nauta2023anecdotal}. Since this definition depends on subjective interpretation, there is no absolute ground truth and thus no direct notion of explanation error \cite{hedstrom2023meta}. Furthermore, explanation quality also depends on the model itself: if the model's reasoning is flawed, even faithful explanations may appear inconsistent or counterintuitive \cite{gilpin_explaining_2018}.  

To address these challenges, different metrics have been proposed across different categories, each capturing distinct aspects of explanation quality. We consider five categories of explanation metrics: \textit{Faithfulness}, \textit{Robustness}, \textit{Localization}, \textit{Complexity}, and \textit{Randomization}. Formally, explanation quality is measured by a metric $\Psi$ evaluating the effectiveness of a method $\Phi$:  
\begin{equation}
    \Psi(\Phi, \boldsymbol{x}, f, \hat{\boldsymbol{y}}) = \hat{q},
\end{equation}
where $\hat{q}$ is a scalar quality score. Explicit parameterization is omitted unless required.

\subsection{Faithfulness}
\label{sec:faithfulness}
\subsubsection{Background}
Faithfulness measures how well explanations reflect the decision process of the model. A faithful explanation identifies features that truly influence predictions. 
Given input $\boldsymbol{x}$ and a set of pixels $V \subseteq [H]\times[W]$ to remove, the perturbation function $\mathcal{P}_{\mathbb{X}}(\boldsymbol{x},V)$ replaces the pixel in $V$ with a baseline value. 
Perturbations are applied either to the most relevant features first (\as{morf}), the least relevant first (\as{lerf}), or randomly.  
We consider two faithfulness metrics: i) \af{fe} \cite{alvarez_melis_towards_2018}, which measures the correlation between feature importance and prediction change, and ii) \af{irof} \cite{rieger_irof_2020}, which evaluates prediction certainty decay when removing relevant segments.


\as{fe} \cite{alvarez_melis_towards_2018} measures the correlation between summed attributions on top-$K$ pixels and the prediction drop when those pixels are removed. Formally,  we define the \as{fe} metric $\Psi_{\mathrm{FE}}$ as:
\begin{equation}
\Psi_{\mathrm{FE}}
=
\mathrm{corr}_{V : |V| = K}
\Bigl(
  \sum_{(j,k)\in V}\Phi(\boldsymbol{x})_{j,k},
  \,f(\boldsymbol{x}) - f\bigl(\mathcal{P}_{\mathbb{X}}(\boldsymbol{x},V)\bigr)
\Bigr).
\end{equation}
Here, $\Phi(\boldsymbol{x})_{j,k}$ denotes the attributed relevancy at pixel $(j,k)$, and $V\subseteq\{1,\dots,H\}\times\{1,\dots,W\}$ is the set of the $K$ pixel indices with the largest attribution values and $\mathrm{corr}$ denotes the Pearson correlation coefficient. A high $\Psi_{\mathrm{FE}}$ value indicates that removing highly attributed pixels substantially lowers the prediction certainty.

\phantomsection \label{metrics:irof}
\as{irof} \cite{rieger_irof_2020} evaluates prediction certainty decay when removing the $K$ most relevant segments rather than pixels. Let $\{S_m\}_{m=1}^M$ be a partition of $\boldsymbol{x}$ into $M$ disjoint segments, which are ranked by their average attribution score. For $\kappa=0,\dots,K$, define the cumulative removal set $V_\kappa=\bigcup_{m=1}^\kappa S_m$ and perturbed input $\mathcal{P}_{\mathbb{X}}(\boldsymbol{x},V_\kappa)$. The IROF metric $\Psi_{\mathrm{IROF}}$ is
\begin{equation}
\Psi_{\mathrm{IROF}}(\boldsymbol{x})
= \frac{1}{K+1}
\sum_{\kappa=0}^{K}
\frac{f\bigl(\mathcal{P}_{\mathbb{X}}(\boldsymbol{x},V_\kappa)\bigr)}
     {f(\boldsymbol{x})}.
\end{equation}
A higher  $\Psi_{\mathrm{IROF}}$ indicates that removing highly attributed segments causes a rapid drop in prediction confidence, and thus reflects a more faithful explanation.

\subsubsection{Limitations}
\label{phantomsec:faithfulness_metrics_problems} Faithfulness metrics inherit several limitations from perturbation-based explanation methods, as both rely on input perturbations to evaluate changes in the prediction certainty. Thus, they face similar issues to those in Section~\ref{subsubsec:perturbation_method_problems}, notably the need for carefully chosen baselines to avoid distribution shifts affecting model predictions.  
In addition, the FE metric relies on the assumption that removing a small number of highly attributed pixels will lead to a measurable change in model output. This assumption implicitly presumes that class-relevant information is concentrated in a small subset of pixels. 
However, \as{rs} scenes often contain multiple spatially disjoint regions associated with the same LULC class. These regions exhibit redundancy, as similar class-relevant features are repeated across the scene.
In such cases, removing a small number of pixels does not produce a meaningful change in prediction, and the attribution sum remains nearly constant across different pixel subsets. As a result, the Pearson correlation used by FE may become numerically unstable, since both numerator and denominator can approach zero, leading to undefined or misleading scores even for faithful explanations.

\begin{figure*}[htp]
\centering
\begin{tabular}{ccccc}
\begin{subfigure}[t]{0.15\textwidth}
    \centering
    \includegraphics[width=\textwidth]{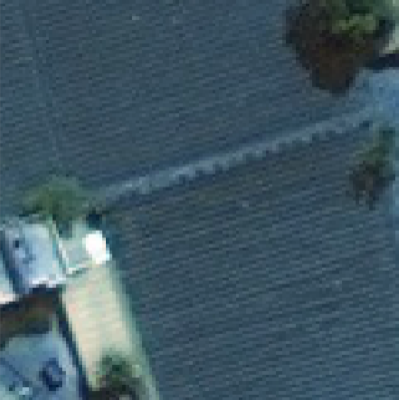}
    \caption{}
    \label{fig:subfig_a}
\end{subfigure} &
\begin{subfigure}[t]{0.15\textwidth}
    \centering
    \includegraphics[width=\textwidth]{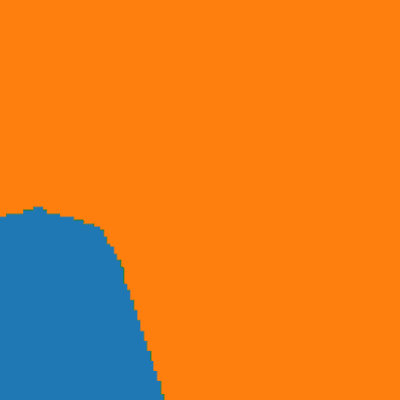}
    \caption{}
    \label{fig:subfig_b}
\end{subfigure} &
\begin{subfigure}[t]{0.15\textwidth}
    \centering
    \includegraphics[width=\textwidth]{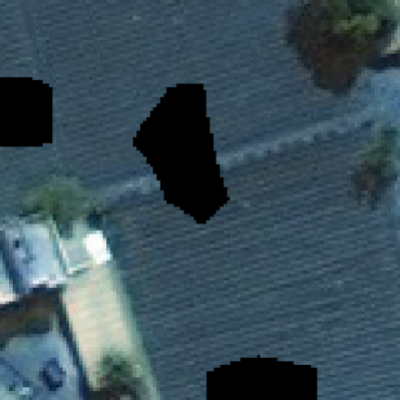}
    \caption{}
    \label{fig:subfig_d}
\end{subfigure} &
\begin{subfigure}[t]{0.15\textwidth}
    \centering
    \includegraphics[width=\textwidth]{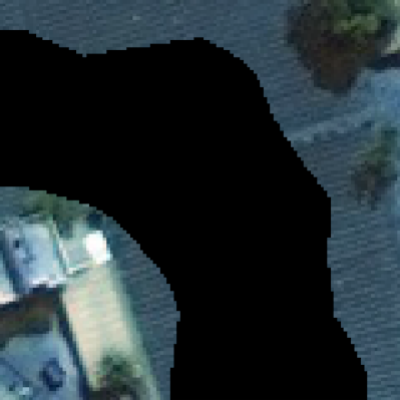}
    \caption{}
    \label{fig:subfig_e}
\end{subfigure} &
\begin{subfigure}[t]{0.15\textwidth}
    \centering
    \includegraphics[width=\textwidth]{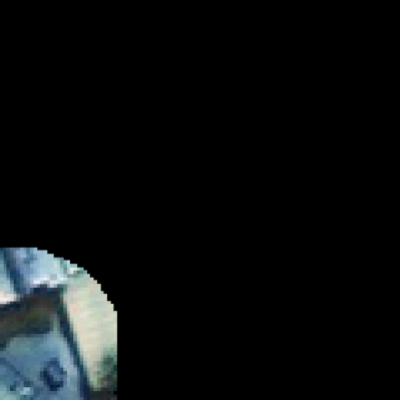}
    \caption{}
    \label{fig:subfig_f}
\end{subfigure} \\

\begin{subfigure}[t]{0.15\textwidth}
    \centering
    \phantom{}%
    \caption*{} 
\end{subfigure} &
\begin{subfigure}[t]{0.15\textwidth}
    \centering
    \includegraphics[width=\textwidth]{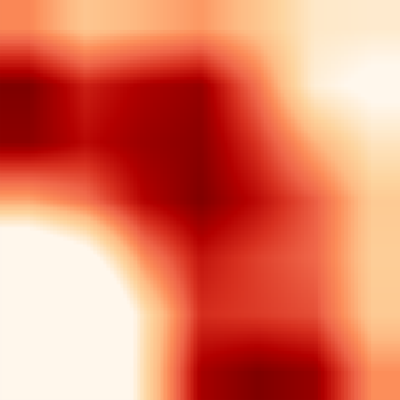}
    \caption{}
    \label{fig:subfig_c}
\end{subfigure} &
\begin{subfigure}[t]{0.15\textwidth}
    \centering
    \includegraphics[width=\textwidth]{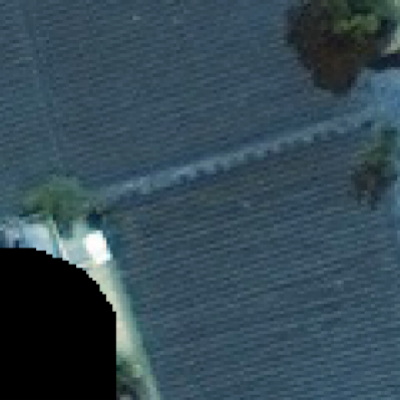}
    \caption{}
    \label{fig:subfig_g}
\end{subfigure} &
\begin{subfigure}[t]{0.15\textwidth}
    \centering
    \includegraphics[width=\textwidth]{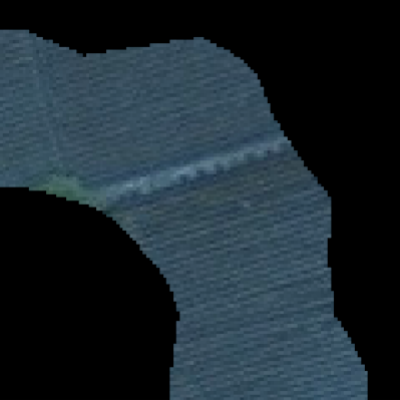}
    \caption{}
    \label{fig:subfig_h}
\end{subfigure} &
\begin{subfigure}[t]{0.15\textwidth}
    \centering
    \includegraphics[width=\textwidth]{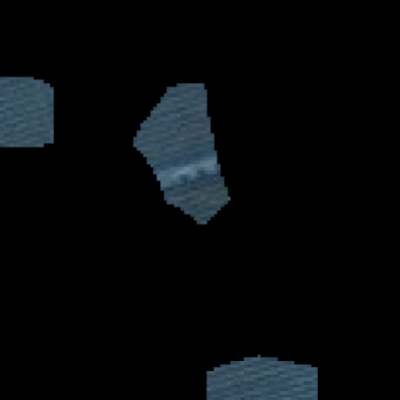}
    \caption{}
    \label{fig:subfig_i}
\end{subfigure} \\
\end{tabular}

\caption{Comparison of \as{morf} and \as{lerf} removal strategies for the target class $c$ (\textit{agricultural land}, orange). Here, $K$ is the percentage of removed pixels and $f_c(\boldsymbol{x})$ represents the prediction certainty.
Top row: 
a) Original image from the \as{dg} \cite{DeepGlobe18} dataset;
b) Pixel-wise reference map. \textit{Agriculture land} in orange;
\as{morf} removal:
c) $K=10\%$, $f_c(\boldsymbol{x})=0.98$;
d) $K=50\%$, $f_c(\boldsymbol{x})=0.64$;
e) $K=90\%$, $f_c(\boldsymbol{x})=0.31$.
Bottom row: 
f) Grad-CAM attribution map for $c$;
\as{lerf} removal:
g) $K=10\%$, $f_c(\boldsymbol{x})=0.99$;
h) $K=50\%$, $f_c(\boldsymbol{x})=0.99$;
i) $K=90\%$, $f_c(\boldsymbol{x})=0.72$.
}
\label{fig:lerf_vs_morf}
\end{figure*}

In contrast to FE,  trajectory-based metrics like IROF assess model behavior across progressively perturbed inputs. Their outcome heavily depends on the feature removal order. Let us consider a scene where LULC class $c_{\text{major}}$ covers most of the image and $c_{\text{minor}}$ only a small region. Perturbing $c_{\text{major}}$ pixels has little effect due to redundancy, so removing most relevant pixels first yields a slow decline in $f(\boldsymbol{x})$, as visualized in Fig.~\ref{fig:lerf_vs_morf}. In contrast, perturbing $c_{\text{minor}}$ pixels quickly lowers prediction certainty, biasing results toward faithful explanations for $c_{\text{minor}}$ and against $c_{\text{major}}$. The \as{lerf} strategy, which removes the least relevant pixels first, mitigates this bias by preserving crucial features of $c_{\text{minor}}$ longer, stabilizing prediction certainty and producing more balanced evaluations.

\subsection{Robustness}
\subsubsection{Background}
The second category we consider is robustness, which assesses the stability of explanations under small input perturbations. 
We require small perturbations $\delta$ such that $\left\| \delta\right\|_2<\varepsilon$ and expect $\left\|\mathcal{P}_{\mathbb{X}}(\boldsymbol{x})-\boldsymbol{x}\right\|_2<\varepsilon$ for a small positive constant $\varepsilon$, assuming that the model output approximately stayed the same $f(\boldsymbol{x}) \approx f\left(\mathcal{P}_{\mathbb{X}}(\boldsymbol{x})\right)$. 
%
To evaluate the robustness of explanation methods for \as{rs} image scene classification, we use i) \af{as} \cite{yeh_fidelity_2019}, which measures sensitivity to small perturbations using a Monte Carlo-based approximation and ii) \af{lle} \cite{alvarez_melis_towards_2018}, which estimates Lipschitz continuity to assess explanation stability under minor input changes.


\as{as} \cite{yeh_fidelity_2019} evaluates the sensitivity of the explanation method by perturbing samples using a Monte-Carlo-based approximation. The \as{as} metric $\Psi_{\mathrm{AS}}$ is defined as:
\begin{equation}
\begin{aligned}
    \Psi_{\mathrm{AS}}=\text{avg}_{\boldsymbol{x}+\delta \in \mathcal{N}_\epsilon(\boldsymbol{x}) \leq \varepsilon}\left[\frac{\|\Phi(\boldsymbol{x})-\Phi(\boldsymbol{x}+\delta)\|}{\|\boldsymbol{x}\|}\right],
\end{aligned}
\end{equation}
where $\varepsilon$ defines the radius of a discrete, finite-sample neighborhood around each input sample $\boldsymbol{x}$. This neighborhood, denoted as $\mathcal{N}_\epsilon(\boldsymbol{x})$, includes all samples that are within a distance of $\varepsilon$ from $\boldsymbol{x}$. A lower score is indicative of more robustness.

The \as{lle} \cite{alvarez_melis_towards_2018} estimates the Lipschitz continuity of the explanation, which measures how much the explanation changes with respect to the input under a minor perturbation, meaning the prediction did not change. The \as{lle} metric $\Psi_{\mathrm{LLE}}$ is defined as:
\begin{equation}
\Psi_{\mathrm{LLE}} = \max_{\boldsymbol{x}+ \delta \in \mathcal{N}_\epsilon(\boldsymbol{x}) \leq \epsilon} \left[\frac{|\Phi(\boldsymbol{x}) - \Phi(\boldsymbol{x}+ \delta)|_2}{|\boldsymbol{x}- (\boldsymbol{x} + \delta)|_2}\right],
\end{equation}
where, $\delta$ is the perturbation vector with $\delta \in \mathcal{N}_\epsilon(\boldsymbol{x}) \leq \epsilon$, and $\mathcal{N}\varepsilon(\boldsymbol{x})$ is the neighborhood of perturbed samples within radius $\varepsilon$. Here, lower values indicate less change concerning the change in input. 

\subsubsection{Limitations}
\label{subsub:robustness_limit}

A key hyperparameter in this metric category is the noise type. If the introduced noise does not naturally occur in the acquired images, it may generate \as{ood} samples, reducing the reliability of the model and thereby affecting the validity of the explanations. When evaluating the robustness of explanation methods, it is important to introduce noise that does not result in \as{ood} samples. This consideration also depends on the model architecture, as many models exhibit inherent resilience to certain types of noise. For instance, in  \as{sar} images, multiplicative noise such as speckle is pervasive \cite{vijay2012speckle}. In contrast, for multispectral images, noise is typically independent of the data and generally additive in nature, such as additive white Gaussian noise \cite{boonprong2018classification, al2010comparative}.

\subsection{Localization}
\subsubsection{Background}
The third category we consider is localization, which measures the spatial accuracy of explanations by comparing them to reference maps or bounding boxes. This comparison can be quantified using a similarity measure $\text{SIM}$ between the generated explanation map $\Phi(\boldsymbol{x})$ and a pixel-level reference map $\boldsymbol{s}_{gt}$. 
%
where $\text{SIM}$ can be a measure such as Intersection over Union, Pearson correlation, or \as{ssim} depending on the nature of the task and reference map.
We utilize i) \af{tki} \cite{theiner_interpretable_2022}, which measures the fraction of top-ranked attributions within the pixel-level reference map, and ii) \af{rra} \cite{arras2022clevr}, which assesses the concentration of high-relevance values within the reference map.

The \as{tki} metric \cite{theiner_interpretable_2022} measures the fraction of the $K$ highest ranked attributions that lie on the binary reference mask $\boldsymbol{s}_{gt}$.  
We define $I_K = \{\,i_1, i_2, \ldots, i_K\}$ and $S_{gt} = \{\,p \mid \boldsymbol{s}_{gt}(p)=1\}$, where $I_K$ is the set of the top-$K$ ranked pixel indices and $S_{gt}$ the set of true-positive pixels in the mask.
The intersection of these sets is $  \boldsymbol{s}_{1-K}  =  I_K  \;\cap\;  S_{gt}$. The \as{tki} metric $\Psi_{\mathrm{TKI}}$ is then defined as:
\begin{equation}
  \Psi_{\mathrm{TKI}}
  =
  \frac{\bigl|\boldsymbol{s}_{1-K}\bigr|}{K}
  .
\end{equation}
If all top-$K$ pixels are within the reference mask $S_{gt}$, then $\Psi_{\mathrm{TKI}} = 1$, therefore, a higher score is desired.

The \as{rra} metric \cite{arras2022clevr} measures the concentration of high-intensity relevance values within the pixel-level reference map. Let $|S_{\text{tot}}|$ be the size of the reference mask. Let us assume that $K$ pixels with the highest relevance values are selected from $\Phi(\boldsymbol{x})$. We then count how many of these selected pixels lie within the reference pixels and divide this number by $|S_{\text{tot}}|$:
\begin{equation}
\Psi_{\mathrm{RRA}}=\frac{|P_{\text{top } K} \cap S_{\text{tot}}|}{|S_{\text{tot}}|}.
\end{equation}
A higher $\Psi_{\mathrm{RRA}}$ indicates stronger localization, as more highly relevant pixels fall within the reference region.

\subsubsection{Limitations}
\label{phansec:limitations_localization}
In \as{rs} scene classification, label dependencies arise when the presence of one label influences the likelihood of another, e.g., ships typically co-occur with sea. Localization metrics do not account for such dependencies. Their reliability also depends on accurate pixel-level reference maps, which are costly to obtain and prone to errors. Outdated or inaccurate annotations compromise the validity of these metrics.  

Additionally, most localization metrics become unreliable when the target class spans a large area of the input image. Let us consider the \as{tki} metric in the scenario where we have a single label for the input image.  Here, the binary reference mask $\mathbf{s}_{gt}$ for the target label becomes a matrix of ones, encompassing all pixels. Thus, $\Psi_{\mathrm{TK}}$ inherently achieves its maximum value of 1, regardless of the explanation $\boldsymbol{e}$. Since $\mathbf{s}_{gt}$ includes every pixel, the intersection $\mathbf{s}_{1-K} = \mathbf{r}_{1-K} \cap \mathbf{s}_{gt}$ reduces to $\mathbf{r}_{1-K}$ itself, as all top-ranked indices in $\mathbf{r}_{1-K}$ necessarily lie within $\mathbf{s}_{gt}$. Consequently, the numerator and denominator of $\Psi_{\mathrm{TK}}$ both equal $K$, yielding $\Psi_{\mathrm{TK}} = 1$. In this case, the metric cannot differentiate meaningful explanations from arbitrary ones.  
This phenomenon is well-documented for simpler localization metrics \cite{kohlbrenner_towards_2020}, such as the Pointing Game \cite{zhang_top-down_2018}. However, in \as{rs} scene classification, this also creates challenges for more sophisticated metrics, as in most \as{rs} scene classification tasks, as LULC classes tend to span larger areas.
\subsection{Complexity}
\subsubsection{Background}
The fourth category we consider is complexity, which quantifies the simplicity and interpretability of explanations, ensuring that they are useful for human understanding. These metrics evaluate properties such as sparsity, entropy, or other measures of complexity in the explanation map $\Phi(\boldsymbol{x})$. 
%
To evaluate the complexity of explanations for \as{rs} image scene classification, we use i) \af{sp} \cite{chalasani_concise_2020}; and ii) \af{co} \cite{bhatt_evaluating_2020}.

\as{sp} \cite{chalasani_concise_2020} is a method for evaluating the sparsity of explanations and is defined as the Gini index of the explanation. It is calculated by summing the product of the ranks of the input features and their attributions and dividing by the sum of the attributions. The \as{sp} metric $\Psi_{\mathrm{SP}}$ is defined as:
\begin{equation}
    \Psi_{\mathrm{SP}}
    = \frac{
        \displaystyle\sum_{j=1}^{H}\sum_{k=1}^{W}
          \bigl(2\,\rho_{j,k} - D - 1\bigr)\, \Phi(\boldsymbol{x})_{j,k}
    }{
        D(D - 1)\,\displaystyle\sum_{j=1}^{H}\sum_{k=1}^{W} \Phi(\boldsymbol{x})_{j,k}
    },
\end{equation}
where $\hat{\boldsymbol e}_{j,k}$ is the attribution of pixel $\boldsymbol{x}_{j,k}$, $\rho_{j,k}$ is the rank of $ \Phi(\boldsymbol{x})_{j,k}$ among all $D$ attributed input features when sorted in non‐decreasing order, $H$ and $W$ denote the image height and width, i.e. $D = H \times W$. A higher $\Psi_{\mathrm{SP}}$ indicates a sparser explanation.

\as{co} \cite{bhatt_evaluating_2020} is defined using the Shannon entropy calculation, which measures the amount of uncertainty or randomness in the attribution map. It is calculated by summing the product of the probabilities of the attributions and the logarithm of the probabilities of the attributions. The \as{co} metric $ \Psi_{\mathrm{CO}}$ is defined as:
\begin{equation}
    \Psi_{\mathrm{CO}}
    = -\sum_{j=1}^{H}\sum_{k=1}^{W}
        p_{j,k}\,\ln\bigl(p_{j,k}\bigr),
\end{equation}
where
\begin{equation}
    p_{j,k} = \frac{| \Phi(\boldsymbol{x})_{j,k}|}{\sum_{a=1}^{H}\sum_{b=1}^{W}| \Phi(\boldsymbol{x})_{a,b}|}
\end{equation}
is the fractional contribution of attribution $ \Phi(\boldsymbol{x})_{j,k}$ , and a higher $\Psi_{\mathrm{CO}}$ indicates greater uncertainty (complexity). A uniformly distributed attribution would have the highest possible complexity score.
\begin{figure}[!t]
    \centering
    \begin{subfigure}[t]{0.3125\columnwidth}
        \includegraphics[width=\textwidth]{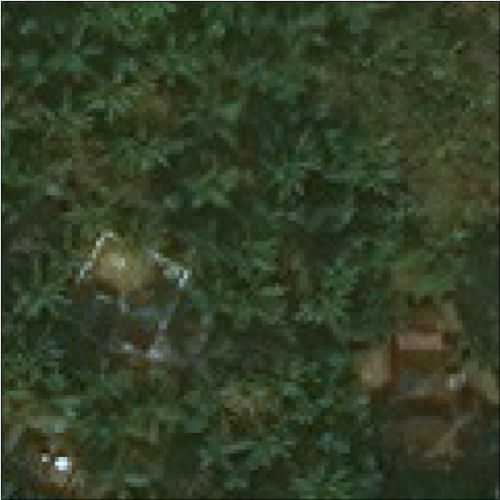}
        \caption{}
    \end{subfigure}
    \begin{subfigure}[t]{0.3125\columnwidth}
        \includegraphics[width=\textwidth]{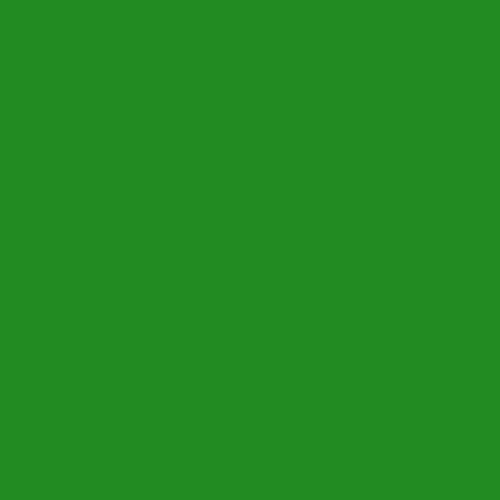}
        \caption{}
    \end{subfigure}
    \par\medskip
    \begin{subfigure}[t]{0.3125\columnwidth}
        \includegraphics[width=\textwidth]{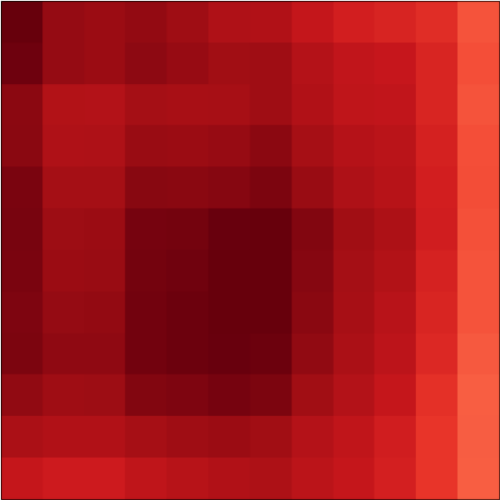}
        \caption{}
        \label{fig:theory_complexity_occ}
    \end{subfigure}
        \begin{subfigure}[t]{0.3125\columnwidth}
        \includegraphics[width=\textwidth]{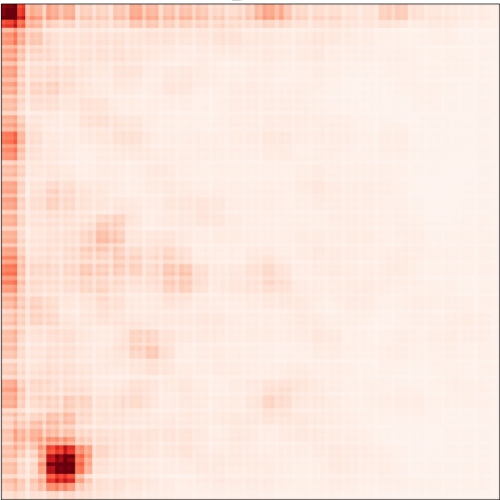}
        \caption{}
        \label{fig:theory_complexity_lrp}
    \end{subfigure}
    \caption{Explanations for an image from the \as{dg} dataset \cite{DeepGlobe18} labeled as Forest and their corresponding \as{sp} metric scores.  
    a) Original image;  
    b) Pixel-wise reference map (green: Forest);  
    c) Occlusion explanation ($\Psi_{\mathrm{SP}} = 0.08$);  
    d) LRP explanation ($\Psi_{\mathrm{SP}} = 0.47$).
    }
    \label{fig:theory_complexity}
\end{figure}
%
\subsubsection{Limitations}
Explanation metrics in the complexity category generally assume that an explanation is simpler if it attributes relevance to fewer features for a given prediction. This assumption originates from natural images, where classes are defined by a few distinctive features, and only a small fraction of pixels contribute to the prediction. However, as discussed before, in \as{rs} scenes LULC classes often cover large areas with repeated class-relevant features, so relevance is distributed across many pixels. Complexity metrics can therefore underestimate explanation quality.
A key limitation of the Gini-based \as{sp} metric $\Psi_{\mathrm{SP}}$ is its tendency to penalize explanations with uniformly distributed attributions. In a scenario where a class spans nearly the entire image, a comprehensive explanation assigns relevance to multiple regions exhibiting discriminative spectral or textural signals. This uniform distribution is interpreted by $\Psi_{\mathrm{SP}}$ as high complexity, thereby misclassifying an accurate explanation as less concise. In contrast, an explanation for a minor class confined to a single region is favored with a low complexity score, even though such concentration does not necessarily improve interpretability.
This limitation is illustrated in Fig.~\ref{fig:theory_complexity}, which shows two explanations for an image labeled as forest, a class that spans nearly the entire image.
The attribution map from the \as{occlusion} method (Fig.~\ref{fig:theory_complexity_occ}) uniformly highlights all forest pixels, yielding a low \as{sp} value ($\Psi_{\mathrm{SP}} = 0.08$). Although most of the input pixels are attributed as relevant, the map remains easy to interpret. On the contrary, the LRP map (Fig.~\ref{fig:theory_complexity_lrp}) concentrates relevance on a small, localized area, producing a higher sparsity score ($\Psi_{\mathrm{SP}} = 0.47$). Here, the \as{sp} metric favors a compact but misleading attribution, mistaking spatial concentration for interpretive simplicity.
%

\subsection{Randomization}
\subsubsection{Background}
The fifth category we consider is randomization, which tests the reliability of explanations by introducing randomness into model parameters or prediction targets. These metrics evaluate the consistency of explanations under such perturbations to ensure that explanations are not artifacts of specific parameter settings or configurations. 
%
Here, we use i) \af{mprt} \cite{hedstrom2024sanity}, which measures explanation complexity under model parameter randomization, and ii) \af{rl} \cite{sixt_when_2020}, which evaluates stability by comparing explanations for target and non-target classes using \af{ssim}.

\as{mprt} \cite{hedstrom2024sanity} measures the relative increase in the complexity of the explanation $\Phi\bigl(\boldsymbol{x}, f_{\mathcal{R}}\bigr)$ from a fully randomized model $f_{\mathcal{R}}$ for image $\boldsymbol{x}$. The \as{mprt} metric $\Psi_{MPRT}$ is defined as:

\begin{equation}
\Psi_{MPRT}
= \frac{
    \xi\bigl(\Phi\bigl(\boldsymbol{x}, f_{\mathcal{R}}\bigr)\bigr)
  - \xi\bigl(\Phi\bigl(\boldsymbol{x}, f\bigr)\bigr)
  }{
    \xi\bigl(\Phi\bigl(\boldsymbol{x},f\bigr)\bigr)
  }.
\end{equation}

Here, $\xi:\mathbb{R}^D\!\to\!\mathbb{R}$ denotes a complexity function. To define $\xi$, a histogram‐entropy measure grounded in Shannon entropy is employed. Specifically, the vectorized attributions $
\boldsymbol{e} = \{\,a_{j,k}\}_{j,k} $ 
are placed into $B$ distinct bins. For the $d$-th bin, the frequency $\eta_d$ is computed and the corresponding normalized probability $\pi_d$ is

\begin{equation}
\xi(\boldsymbol{e})
= -\sum_{d=1}^B \pi_d \,\log\!\bigl(\pi_d\bigr),
\quad\text{where}\quad
\pi_d = \frac{\eta_d}{\sum_{d'=1}^B \eta_{d'}}\,.
\end{equation}

The \as{rl} metric \cite{sixt_when_2020} is defined as the \as{ssim} over the explanation of the reference label and an explanation of the non-target class $\boldsymbol{y}^{\prime}$. The \as{rs} metric $\Psi_{\mathrm{RL}}$ is defined as:
\begin{equation}
    \Psi_{\mathrm{RL}}
    =\operatorname{SSIM}\left(\Phi(\boldsymbol{x}, f, \hat{\boldsymbol{y}}), \Phi\left(\boldsymbol{x}, f, \boldsymbol{y}^{\prime}\right)\right),
\end{equation}
where $\Phi(\boldsymbol{x}, f, \hat{\boldsymbol{y}})$ is the explanation generated for the prediction $\hat{\boldsymbol{y}}$ and $\Phi\left(\boldsymbol{x}, f, \boldsymbol{y}^{\prime}\right)$ is the explanation generated for a non-target class $\boldsymbol{y}^{\prime}$. Lower values indicate that the explanations are not correlated.

\subsubsection{Limitations}
Due to their minimal sensitivity to variations in input data characteristics, randomization metrics are also robust for \as{rs} image classification. Metrics like \as{mprt} focus solely on how the randomization of model weights affects the outcomes of the explanation method, isolating the influence of the parameters the model has learned.
In contrast, \as{rl} measures the similarity between explanations for a target class and a non-target class. However, this metric may encounter challenges with label dependencies. For example, if the target class and the non-target class are both influenced by a shared present class, regions associated with the shared class may exhibit high relevance for both classes. In such cases, the attribution similarity can be disproportionately high for both target and non-target classes, because the shared class dominates the explanation outcomes.

\section{Dataset Description and Experimental Design}
\label{sec:data_experiments}
\subsection{Dataset Description}

The experiments are conducted using three datasets: 1) \af{ben} \cite{clasen2024reben}, 2) \af{dg} \cite{DeepGlobe18, burgert_label_2023},  and 3) \af{fbp} \cite{FBP2023}.  Table \ref{table:ds_characteristics} summarizes the key characteristics of the considered datasets.

\begin{table*}[t!]
\caption[Summary of key attributes of the considered \as{rs} datasets.]{Summary of key attributes of utilized \as{rs} datasets, including the number of images ($|D|$), unique labels ($L$), average labels per image, number of bands ($C$), image sizes and the type of pixel-level reference maps.} \label{table:ds_characteristics}
\centering
\renewcommand*{\arraystretch}{1.2}
\sisetup{group-separator={,}, input-ignore={,}}
\begin{tabularx}{\linewidth}{p{3cm} S[table-format=6] S[table-format=2] S[table-format=1.2] S[table-format=2] >{\centering\arraybackslash}X r}
    \toprule
    \multirow{2}{*}{Dataset} 
    & {\multirow{2}{*}{$|\mathcal{D}|$}} 
    & {\multirow{2}{*}{\hspace{1.2em}$L$}\hspace{1.2em}} 
    & {Avg. $L$} 
    & {\multirow{2}{*}{\hspace{1.2em}$C$\hspace{1.2em}}} 
    & {\multirow{2}{*}{Image Size (Spatial Resolution)}} 
    & {\multirow{2}{*}{Pixel-Level Reference Maps}} \\
    & & & {per Image} &  &  &  \\ 
    \midrule
    \as{ben} \cite{clasen2024reben} & 250,249 & 19 & 2.95 & 10 & 120$\times$120 (10\,m) & Thematic Product \\
    \as{dg} \cite{DeepGlobe18} & 18,185 & 6 & 1.71 & 3 & 120$\times$120 (0.5\,m) & Manually Annotated \\ 
    \as{fbp} \cite{FBP2023} & 109,200 & 24 & 2.61 & 4 & 256$\times$256 (4\,m) & Manually Annotated \\
    \bottomrule
\end{tabularx}
\end{table*}


\noindent BigEarthNet-S2 v2.0 (denoted as \as{ben} hereafter)  \cite{clasen2024reben} is a multi-label dataset, including 549\,488  Sentinel-2 images acquired over ten countries in Europe. We used all the bands of the Sentinel-2 images, and thus applied cubic interpolation to 20-m  and 60-m bands. Each image is associated with one or more class labels (i.e., multi-labels) based on the 19 class nomenclature \cite{sumbul2021bigearthnet}. In the experiments, we use the official train-validation-test split defined in \cite{clasen2024reben}.

\noindent \as{dg} is a multi-label dataset, which Burgert et al.\ \cite{burgert_label_2023} derived from the original DeepGlobe Land-Cover Classification Challenge dataset \cite{DeepGlobe18}. The original DeepGlobe corpus comprises 1,949 manually annotated RGB tiles of size $2448 \times 2448$ pixels at 0.5 m spatial resolution, collected over Thailand, Indonesia, and India. Each tile is divided into 400 images, each of which is a section of $120 \times 120$ pixel images. 
Multi-labels were assigned from the corresponding pixel-level reference maps, while images containing the class \enquote{unknown} were discarded. Only 20\% of single-class images and all multi-class images are retained, yielding 30\,443 samples split into training (60\%), validation (20\%), and test (20\%) subsets.

\noindent \as{fbp} is a multi-label dataset derived from the Five Billion Pixels (FBP) collection \cite{FBP2023}, comprising 150 Gaofen-2 RGB–NIR scenes (4 m spatial resolution) that together cover roughly 50 000 km$^{2}$ in China. Following the procedure used for \as{dg}, each tile is partitioned into an area of $256 \times 256$-pixel images and labeled with a 24-class scheme that distinguishes artificial, agricultural, and natural surfaces. The final dataset consists of a training set with 87\,360 samples (80\%), a validation set with 10\,920 samples (10\%), and a test set with 10\,920 samples (10\%).

\subsection{Evaluation Protocol}
In the experiments, we evaluate explanation methods and metrics for feature attribution in \as{rs} image scene classification using the pipeline illustrated in Fig.~\ref{fig:main-figure}. We generate explanations by applying different feature attribution methods to the prediction of \as{dl} models for the considered \as{rs} datasets. Next, we estimate the quality of these explanations using the considered explanation metrics. Then, the reliability of these metrics is assessed by using the MetaQuantus \cite{hedstrom2023meta} framework. 

As the quality of an explanation cannot be objectively measured due to the lack of ground truth \cite{hedstrom2023meta}, the explanation metrics are tested against two failure modes: \af{nr} of a metric, which is the resilience against minor perturbations, and \af{ar}, which is the reactivity of the metrics to disruptive perturbations. 
A perturbation is considered minor when the predicted class label does not change, and disruptive when the predicted class label changes.
To test against these failure modes, two consistency criteria \af{iac} and \af{iec} are utilized. 


The IAC criterion measures whether the explanation scores remain similarly distributed after minor perturbations and whether they become dissimilarly distributed following disruptive perturbations. Let \(\hat{\mathbf{q}} \in \mathbb{R}^N\) be the unperturbed metric scores for \(N\) samples and \(\mathbf{q}'_k \in \mathbb{R}^N\) the corresponding perturbed scores for \(k = 1,\dots,K\). The IAC score is defined as:
\begin{equation}
\text{IAC} \;=\;\frac{1}{K}\,\sum_{k=1}^K\,d\bigl(\hat{\mathbf{q}}, \mathbf{q}'_k\bigr),
\end{equation}
where \(d\) is the Wilcoxon signed-rank test and \(0 \le \text{IAC} \le 1\). A higher value indicates that unperturbed and perturbed metric scores are more similarly distributed.

The IEC criterion evaluates the consistency of metrics in preserving ranking order across multiple explanation methods. 
Let $ \bar{\mathbf{Q}} \in \mathbb{R}^{N \times L} $ be the unperturbed metric scores for $ N $ samples and $ L $ explanation methods, averaged over $ K $ perturbations.
Additionally, let $ \overline{\mathbf{Q}}' \in \mathbb{R}^{N \times L} $ be the corresponding averaged perturbed metric scores.
We define a binary ranking agreement matrix $ \mathbf{U}^t \in [0,1]^{N \times L} $, where $ t \in \{M, D\} $ indicates a minor ($M$) or disruptive ($D$) perturbation.
The IEC score is defined as:
\begin{equation}
\text{IEC}=\frac{1}{N \times L}\sum_{i=1}^N\sum_{j=1}^LU_{i,j}^t,
\end{equation}
with $0 \le \text{IEC} \le 1$.
The entries $U_{i,j}^t$ capture how consistently the ranking of the metrics agrees between unperturbed $ \bar{\mathbf{Q}} $ and perturbed $ \overline{\mathbf{Q}}' $. 
For minor perturbations ($t=M$) the ordering of the $L$ methods for each sample must remain identical, whereas for disruptive perturbations ($t=D$), the unperturbed score must always be ranked higher than its perturbed counterpart.

The \af{mc} score combines both intra- and inter-consistency criteria across different failure modes as:
\begin{equation}
\mathbf{MC} \;=\; \left( \frac{1}{|\boldsymbol{m}^*|} \right) \boldsymbol{m}^{*T} \boldsymbol{m},
\end{equation}
where
\begin{equation}
\label{eq:mc_score}
\boldsymbol{m} 
\;=\;
\begin{bmatrix}
\mathrm{IAC}_{NR} \\
\mathrm{IAC}_{AR} \\
\mathrm{IEC}_{NR} \\
\mathrm{IEC}_{AR}
\end{bmatrix}
\quad\text{and}\quad
\boldsymbol{m}^* \;=\; \mathbb{1}^4,
\end{equation}
with the all-ones vector representing an ideal explanation metric that achieves a value of 1 on all criteria. A metric with an $\mathbf{MC}$ score close to 1 indicates strong performance across all criteria tested, with higher values reflecting greater overall robustness and reliability.
The $\mathbf{MC}$ score is averaged over two types of perturbation: model perturbation and input perturbation. In model perturbation, random noise is added to the parameters, whereas in input perturbation, it is added to the input data. 

While the protocol provides a systematic assessment of explanation metrics, it is computationally demanding as its runtime depends on the efficiency of the employed explanation methods and metrics. In particular, perturbation-based methods and metrics require multiple forward passes and repeated perturbations, which significantly increases computation time compared to closed-form or gradient-based approaches. This limitation is not inherent to the evaluation protocol but reflects the methodological design of certain metrics. A detailed comparison of computational efficiency across methods and metrics is provided in \hbox{subsection~\ref{subsec:efficiency}.}

\subsection{Experimental Setup}
We have systematically evaluated ten explanation metrics across five widely used feature attribution methods on multi-label \as{rs} datasets (\as{ben}, \as{dg}, and \as{fbp}). To this end, we trained a ResNet-50 model on \as{ben} and ResNet-18 models on \as{dg} and \as{fbp}, subsequently computing explanations for the respective model predictions. 
The ResNet architecture was selected due to its robust performance and demonstrated efficacy when paired with popular explanation methods.

We evaluated the quality of the explanation using the considered metrics and evaluated their reliability with MetaQuantus \cite{hedstrom2023meta} on 512 randomly drawn samples, a sample size demonstrated to be sufficient in previous work \cite{hedstrom2023meta}. The hyperparameters were set as in the original publication ($k=5$, three iterations, which were shown in \mbox{ \cite{hedstrom2023meta} }to be effective across different datasets. Gaussian noise was chosen for the input perturbation to preserve physical consistency, in line with the recommendations in \cite{burgert2024estimating}.
The choice of model depth does not substantially affect the outcome of the evaluation, as the performance of MetaQuantus has been shown to be largely invariant with respect to model architecture and depth \cite{hedstrom2023meta}. For model training, we used the AdamW optimizer, initialized at a learning rate of \(10^{-2}\) and a weight decay of $0.9$. All experiments were performed on Nvidia H100 80GB GPUs.

We limit our analysis to five explanation methods, as prior work \cite{hedstrom2023meta} indicates a minimal influence of the number of methods on the result scores.
Additionally, we used a Random baseline explanation method that samples values randomly from a uniform distribution $U(0,1)$.
For a multi-label case with a label set $\{1,…, K\}$, we apply the same method separately to each label $c \in \{1,…, K\}$ to obtain a feature attribution map for each class $c$. Furthermore, as is common in the literature, we summed the feature attributions over the channels.

The hyperparameters for the explanation methods were configured as follows. For \as{deeplift}, the perturbation baseline value was set to $0$. For \as{gradcam}, the last convolutional layer was analyzed, and the explanations were upscaled by bilinear interpolation. The \as{lime} method employed a Euclidean exponential kernel (kernel width 500) as a similarity measure, used linear regression as an interpretable model, and generated 15 superpixels using \as{slic}. For \as{lrp}, following \cite{montavon_layer-wise_2019}, the first third of the network was initialized with the LRP-$\gamma$ rule, the second third with the LRP-$\epsilon$ rule and the final third with the LRP-$0$ rule. Lastly, the \as{occlusion} method employed a sliding window of size $(50, 50)$ with strides of $(10, 10)$ across the channels, and the perturbation baseline was set to $0$. 
%
The quality of the explanations was assessed using 10 explanation metrics implemented via the Quantus framework \cite{hedstrom2023quantus} with default hyperparameters. Quantus is recognized as the most suitable framework for evaluating explanations \cite{le2023benchmarking}. The explanation metrics were computed over 1024 randomly drawn test samples from each dataset. Both explanation methods and metrics were used with their standard parameters to evaluate the configurations that are most commonly used in \hbox{\acrshort{rs}.}
The explanation methods are implemented using the Captum framework \cite{kokhlikyan2020captum} with modifications to Captum, Quantus, and MetaQuantus to support multi-label image scene classification.

\begin{figure*}[t]
    \centering
    \begin{subfigure}[t]{0.5\textwidth}
        \centering
        \includegraphics[width=\textwidth]{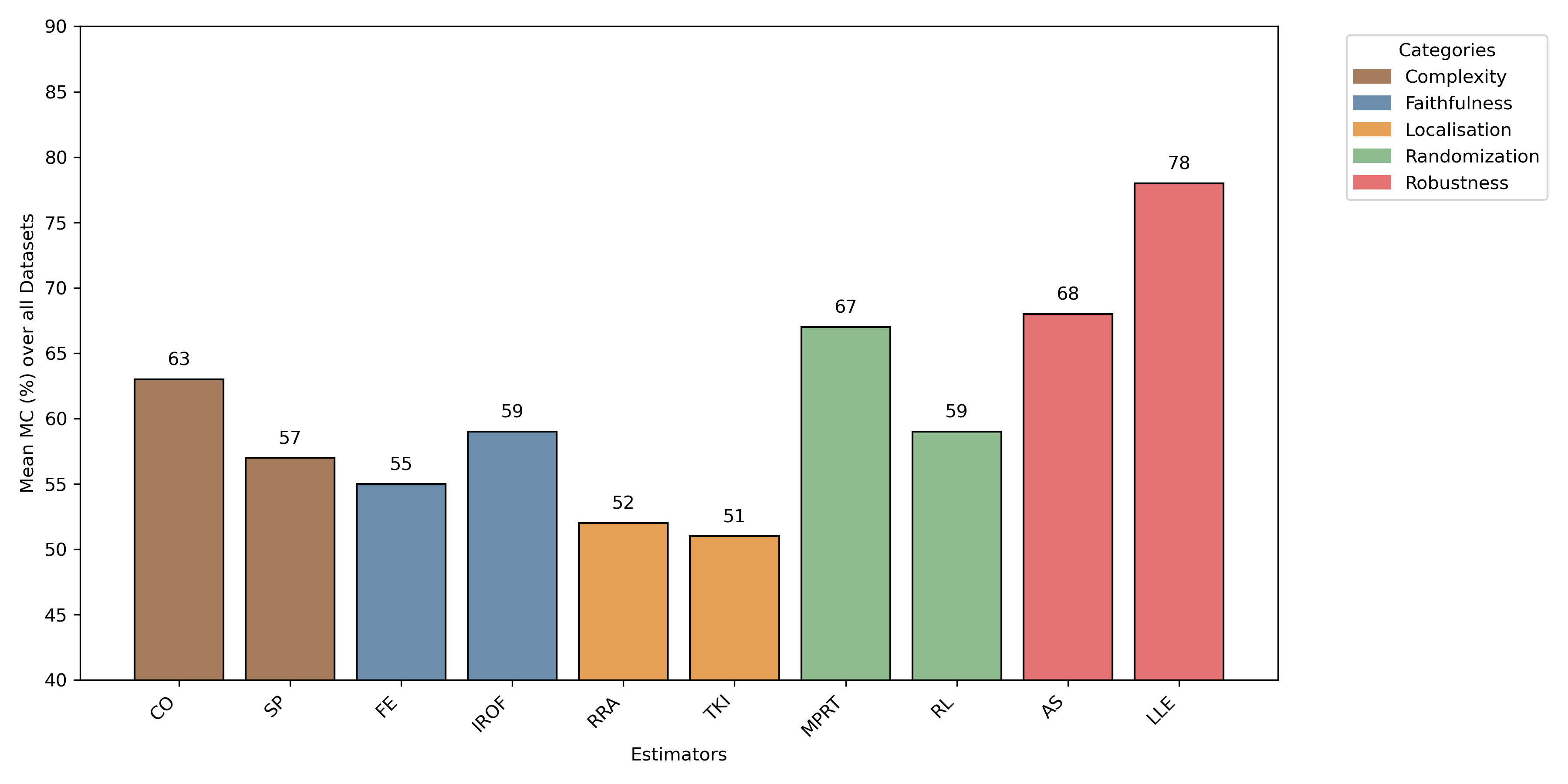}
        \caption{}
        \label{fig:mc_each_dataset_fig1}
    \end{subfigure}%
    ~
    \begin{subfigure}[t]{0.5\textwidth}
        \centering
        \includegraphics[width=\textwidth]{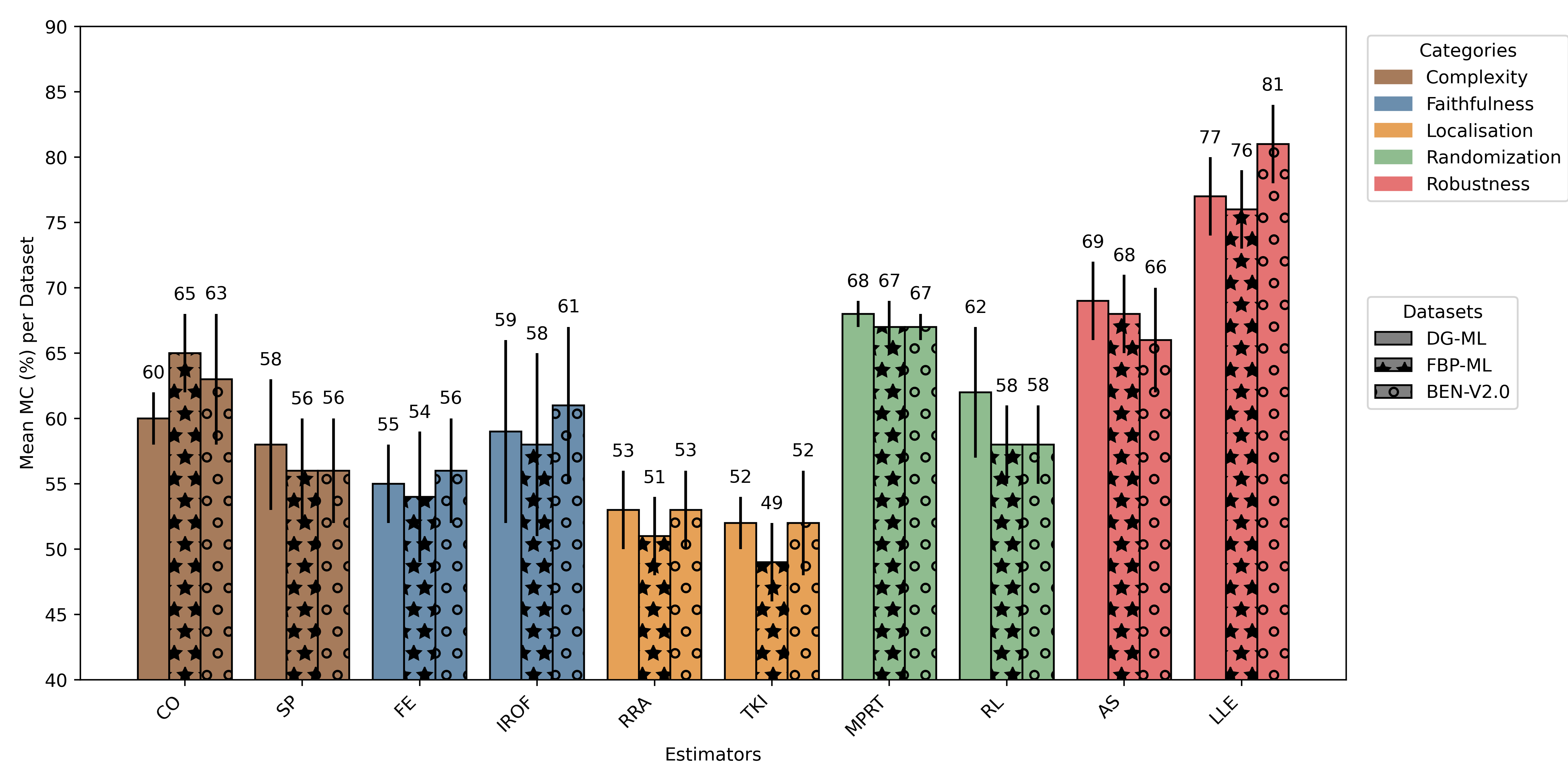}
        \caption{}
        \label{fig:mc_each_dataset_fig2}
    \end{subfigure}
\caption{Mean $\mathbf{MC}$ scores for the considered explanation metrics:
a) scores associated to all considered \as{rs} datasets (\as{dg} \cite{DeepGlobe18}, \as{fbp} \cite{FBP2023} and \as{ben} \cite{clasen2024reben});
b) scores associated to each individual dataset (\as{dg} \cite{DeepGlobe18}, \as{fbp} \cite{FBP2023} and \as{ben} \cite{clasen2024reben}).
Metrics are categorized into five color-coded groups:  i) Complexity (brown), ii) Faithfulness (blue), iii) Localization (orange), iv) Randomization (green), and v) Robustness (red).
}
\label{fig:mc_side_by_side}
\end{figure*}
\section{Experimental Results}
\label{sec:results}
We carried out different kinds of experiments in order to i) assess the reliability of explanation metrics for \as{rs} image scene classification problems, and ii) evaluate the effectiveness of feature attribution methods based on the most reliable explanation metrics.
This ordering ensures that the selection of suitable explanation methods is grounded in metrics whose reliability has been rigorously validated.

\subsection{Evaluation of the considered explanation metrics}

In Fig.~\ref{fig:mc_each_dataset_fig1}, we report the mean MC score of each explanation metric, where the scores have been averaged over all three multi‑label \as{rs} datasets (\as{ben}, \as{dg}, and \as{fbp}), resulting in a single bar per metric. In contrast, Fig.~\ref{fig:mc_each_dataset_fig2} decomposes these results by dataset. For each explanation metric, three separate bars show its mean MC score on \as{dg}, \as{fbp}, and \as{ben} individually. Across both figures, the magnitude of the metrics remains stable, indicating that metric reliability is robust whether computed on the combined dataset or on each dataset in isolation.

From Fig.~\ref{fig:mc_each_dataset_fig1}, one can see that the metrics from the robustness category are the most reliable for \as{rs} image scene classification. In particular, \as{lle} achieves the highest mean MC score of $0.79$, while \as{as} attains the second-highest score of $0.68$. 
This is because the metrics in this category have fewer limitations compared to other categories and are therefore particularly well-suited for \acrshort{rs} image scene classification (see Section~\ref{subsub:robustness_limit}).
By observing the figure, one can also see that the localization metrics, \as{rra} and \as{tki}, perform worst with mean MC scores of  $0.52$ and $0.51$, respectively. The primary cause of the low performance of localization metrics arises from their methodological constraint, which is that they are unreliable when there is a single LULC class present in $\boldsymbol{x}$.

\begin{figure}[h!]
    \centering
    \begin{subfigure}[t]{0.2\textwidth}
        \centering
        \includegraphics[width=\linewidth]{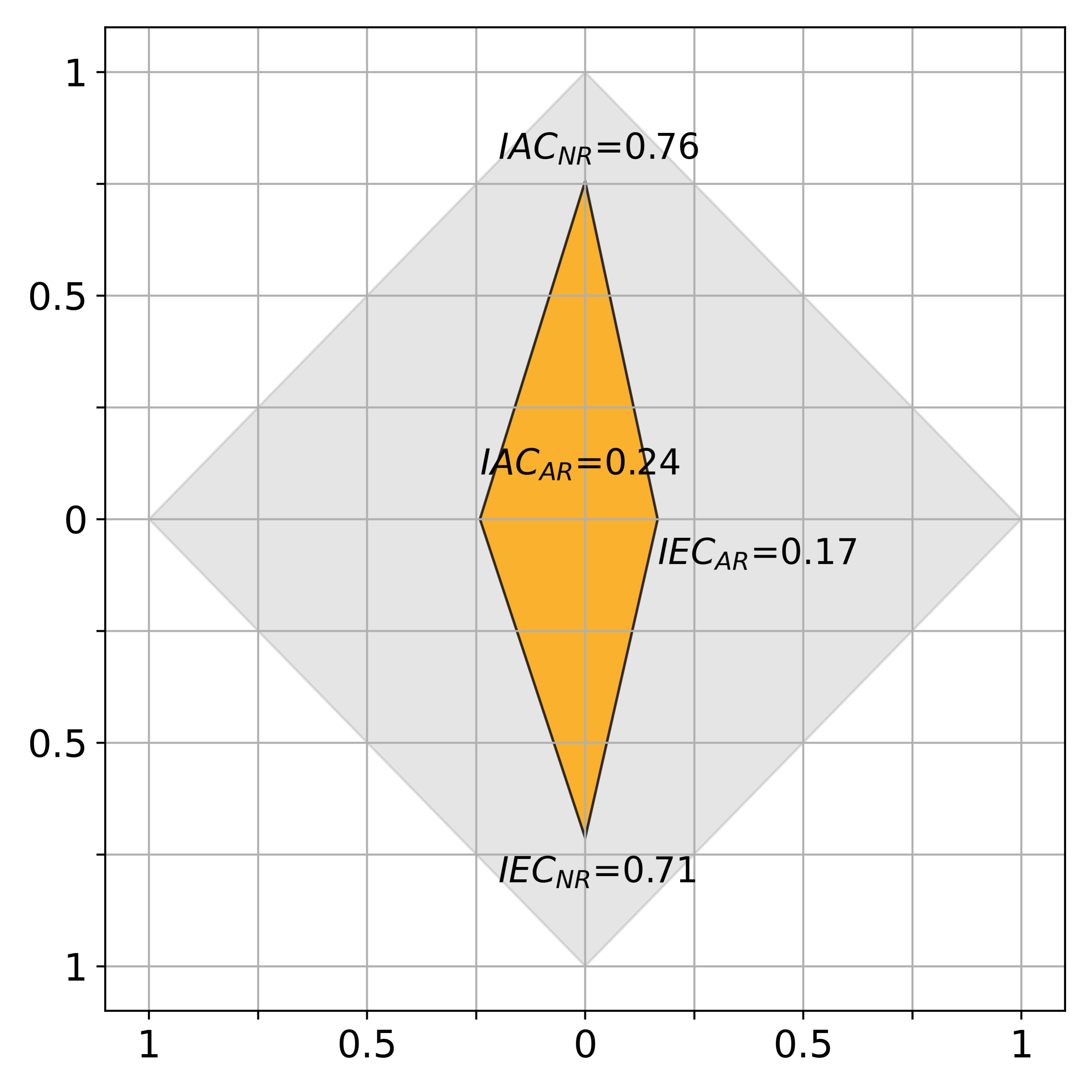}
        \caption{}
        \label{fig:tki}
    \end{subfigure}
    \hspace{0.05\textwidth} 
    \begin{subfigure}[t]{0.2\textwidth}
        \centering
        \includegraphics[width=\linewidth]{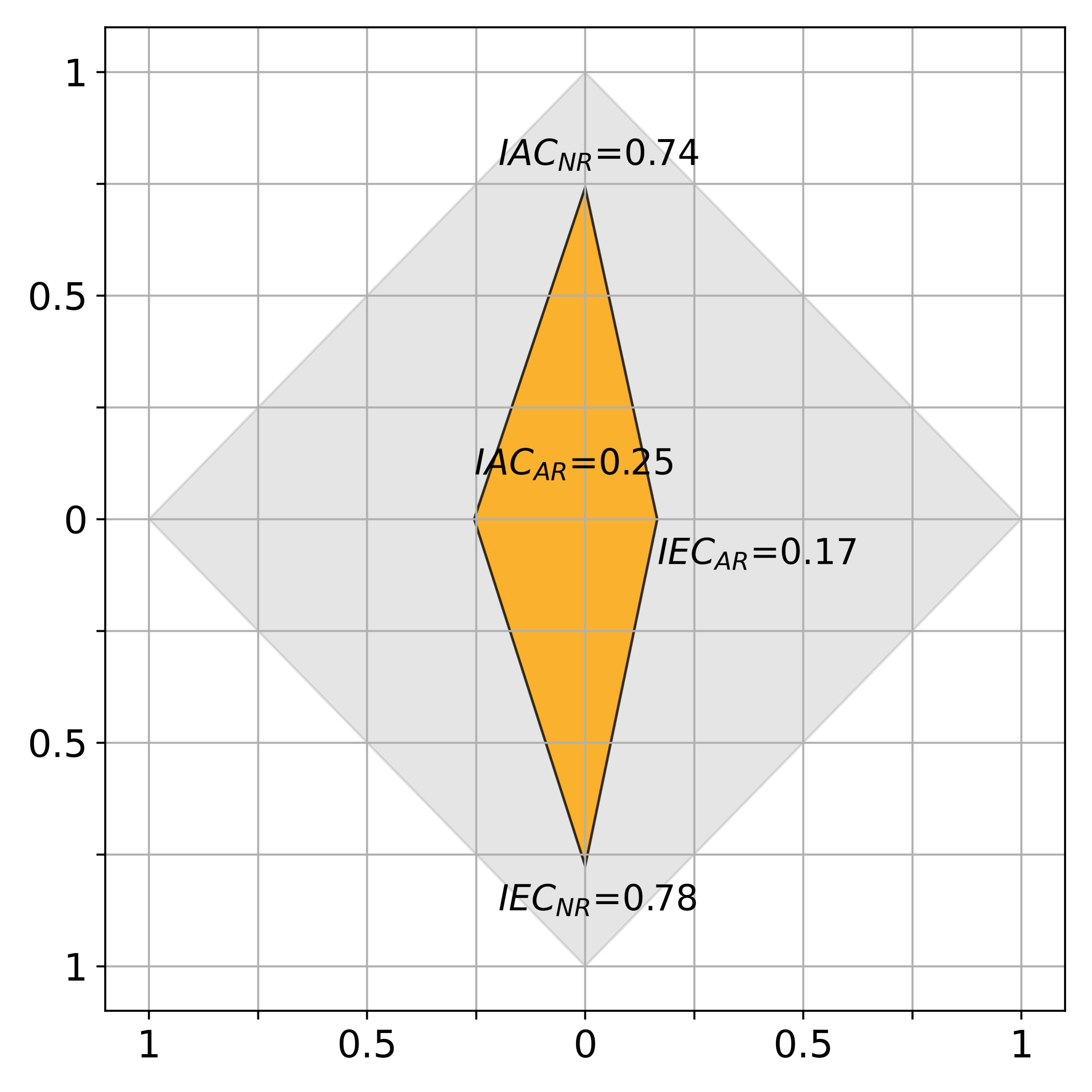}
        \caption{}
        \label{fig:rra}
    \end{subfigure}
    \caption{Results of the evaluation protocol on the \as{dg} dataset \cite{DeepGlobe18} for the localization metrics: a) \textit{\al{tki}} and b) \textit{\al{rra}}.}
    \label{fig:localization_single_area}
\end{figure}

On a closer look, the evaluation of the input perturbation test on these metrics (\as{tki} in Fig.~\ref{fig:tki}  and \as{rra} in Fig.~\ref{fig:rra}) reveals consistent performance patterns. Specifically, the input perturbation test yields similar trends across the four evaluation scores that contribute to the MC score (see Eq.~\ref{eq:mc_score}). Both \as{tki} and \as{rra} exhibit high values for $\text{IAC}_\text{NR}$ and relatively high $\text{IEC}_\text{NR}$ values. On the other hand, both metrics show low values for $\text{IAC}_\text{AR}$ and $\text{IEC}_\text{AR}$.

The NR scores indicate consistency over minor perturbations, while the AR scores reflect changes in response to disruptive perturbations. The pattern observed in these scores responds to the methodological flaws of the localization metrics described in Section~\ref{phansec:limitations_localization}. Specifically, when a single LULC class spans the entire image, the explanation coincides exactly with the reference map. 
Therefore, the metric always yields an output value of $1$. 
Since the  $\text{IAC}_\text{ER}$ and $\text{IEC}_\text{ER}$ scores measure the metrics consistency over minor perturbations, both scores consistently assign the highest score of 1.
However, as the output is constant, disruptive perturbations cannot change the output of the metric, leading to low $\text{IAC}_\text{AR}$ and $\text{IEC}_\text{AR}$ values.
This flaw is particularly visible in the \as{dg} dataset, as it has the lowest average amount of LULC classes per image of the considered datasets.

Furthermore, it is visible in Fig.~\ref{fig:mc_each_dataset_fig1} that the randomization metrics also perform comparatively well with the second-highest average scores. This is in accordance with the assumption that the metrics of this category perform independently of the data. From the figure, one can see that the \as{mprt} metric achieves a mean MC score of $0.68$, while the \as{rl} metric attains a score of $0.60$. 
\begin{figure}[h!]
    \centering
    \begin{subfigure}[t]{0.2\textwidth}
        \centering
        \includegraphics[width=\linewidth]{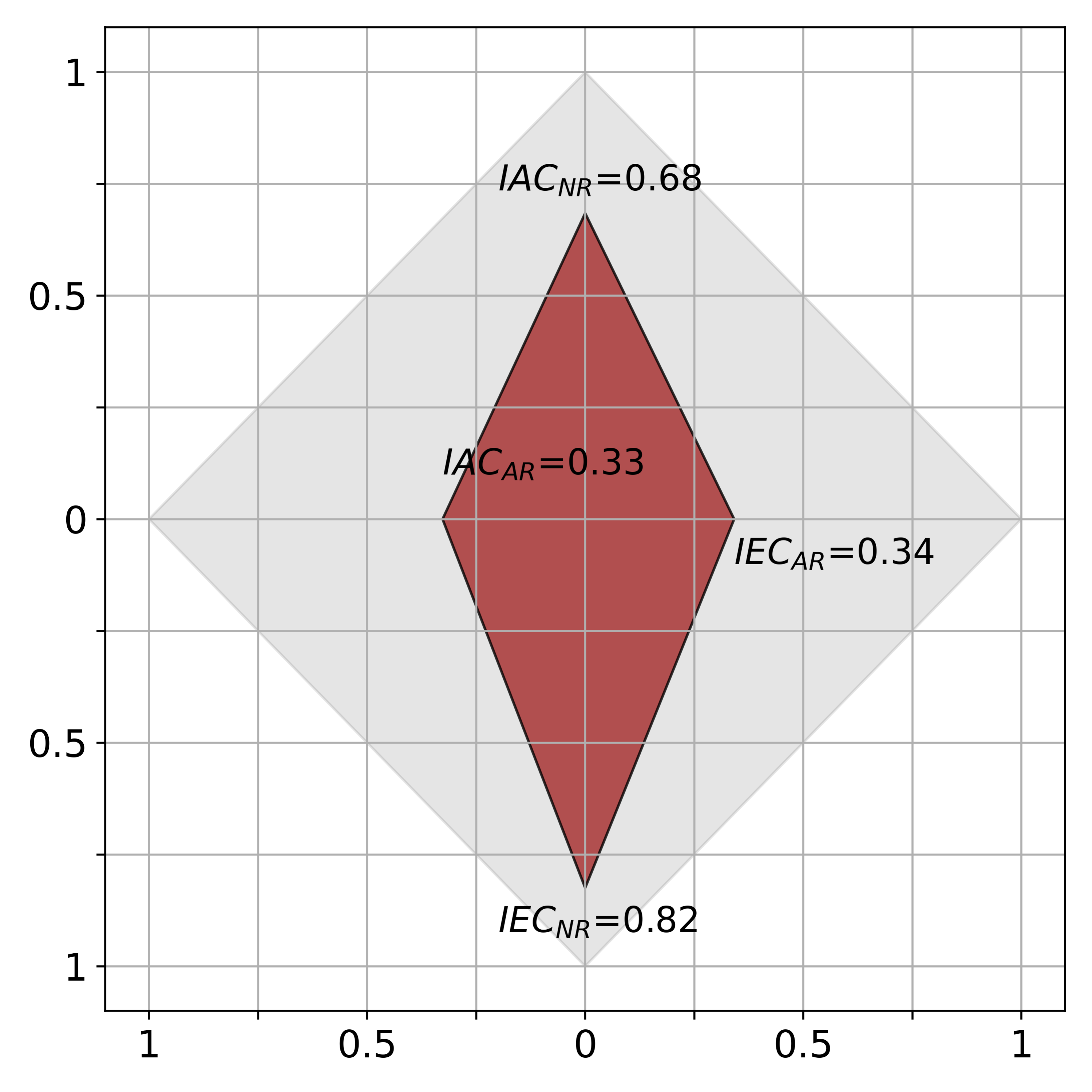}
        \caption{}
        \label{fig:complexity}
    \end{subfigure}
    \hspace{0.05\textwidth} 
    \begin{subfigure}[t]{0.2\textwidth}
        \centering
        \includegraphics[width=\linewidth]{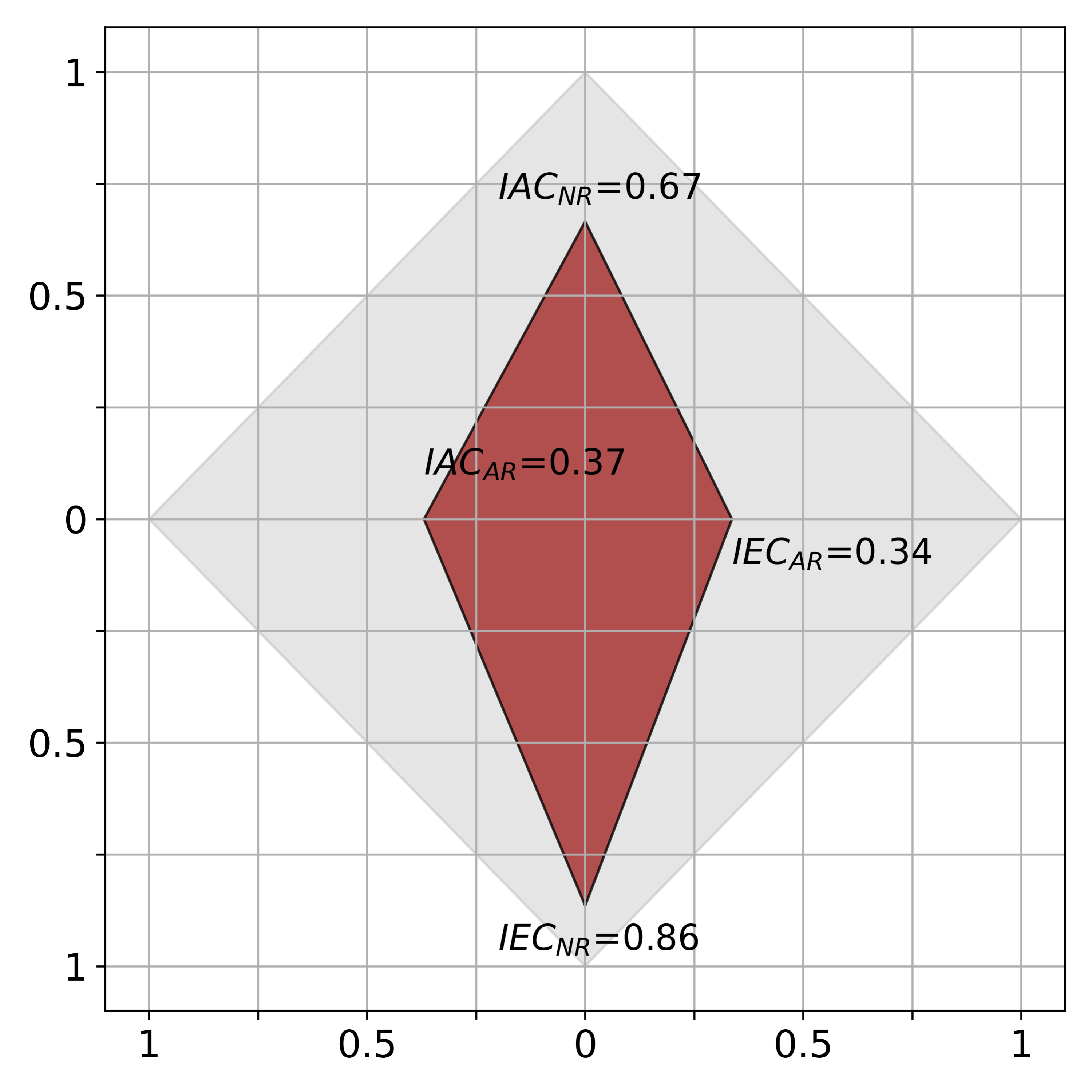}
        \caption{}
        \label{fig:sparseness}
    \end{subfigure}
        \caption{Results of the evaluation protocol on the \as{dg} dataset \cite{DeepGlobe18} for the complexity metrics: a) \textit{\al{co}}; and b) \textit{\al{sp}}.}
    \label{fig:complexity_single_area}
\end{figure}
The mean MC values for the metrics \as{co} and \as{sp} in the complexity category are $0.63$ and $0.57$, as visualized in Fig.~\ref{fig:mc_each_dataset_fig1}. 
Similar to the localization metrics, the complexity metrics also struggle with the characteristic of \as{rs} image data. Single LULC classes can span large parts of an image, yet only a small region provides the discriminative information necessary for accurate classification.
In this case, the metrics tend to attribute higher complexity to the explanation.
This deficit is again reflected in the low $\text{IAC}_\text{AR}$ and $\text{IEC}_\text{AR}$ values, as shown for the \as{co} metric in Fig.~\ref{fig:complexity} and the \as{sp} metric in Fig.~\ref{fig:sparseness}.

\begin{figure}[h!]
    \centering
    \begin{subfigure}[t]{0.2\textwidth}
        \centering
        \includegraphics[width=\linewidth]{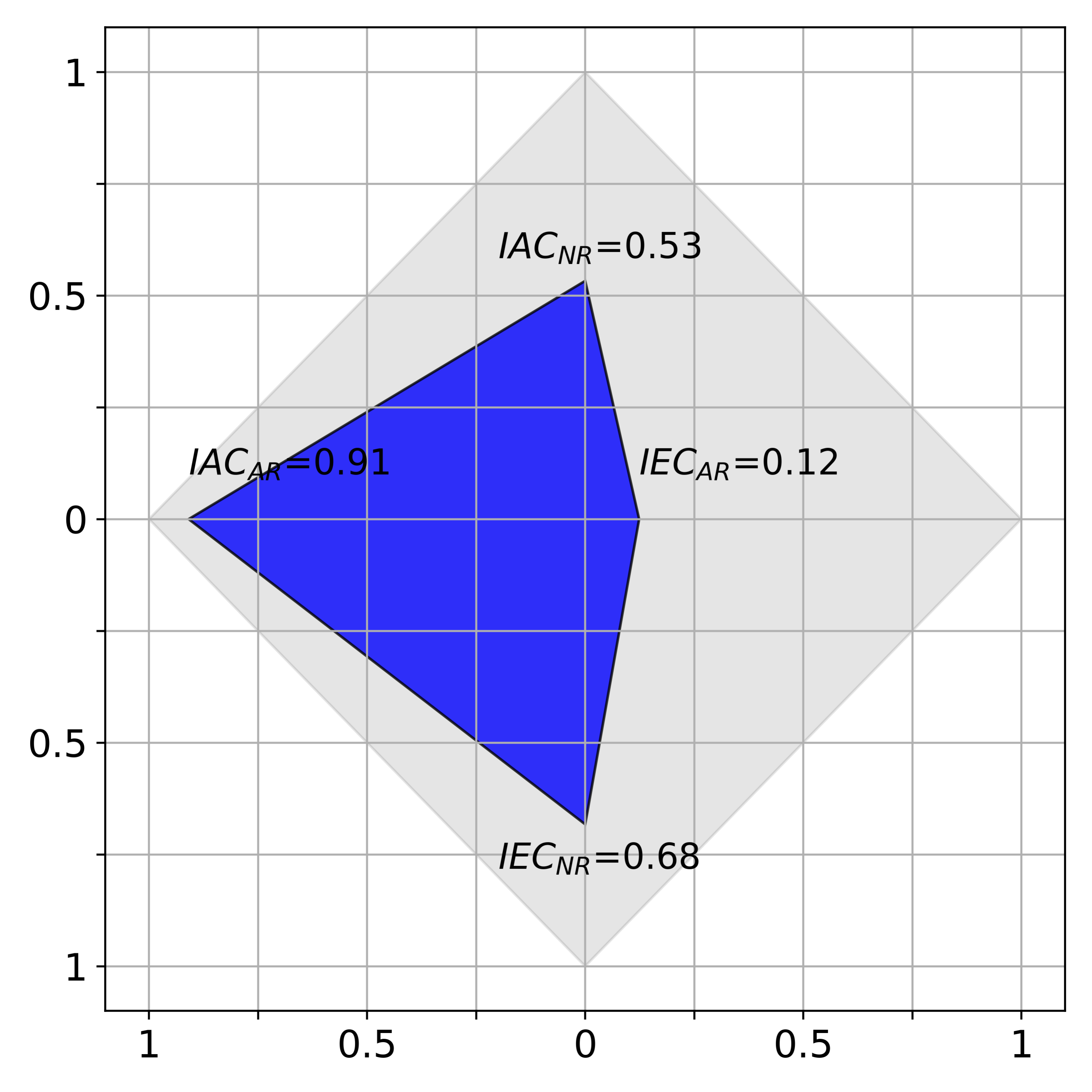}
        \caption{}
        \label{fig:fe}
    \end{subfigure}
    \hspace{0.05\textwidth} 
    \begin{subfigure}[t]{0.2\textwidth}
        \centering
        \includegraphics[width=\linewidth]{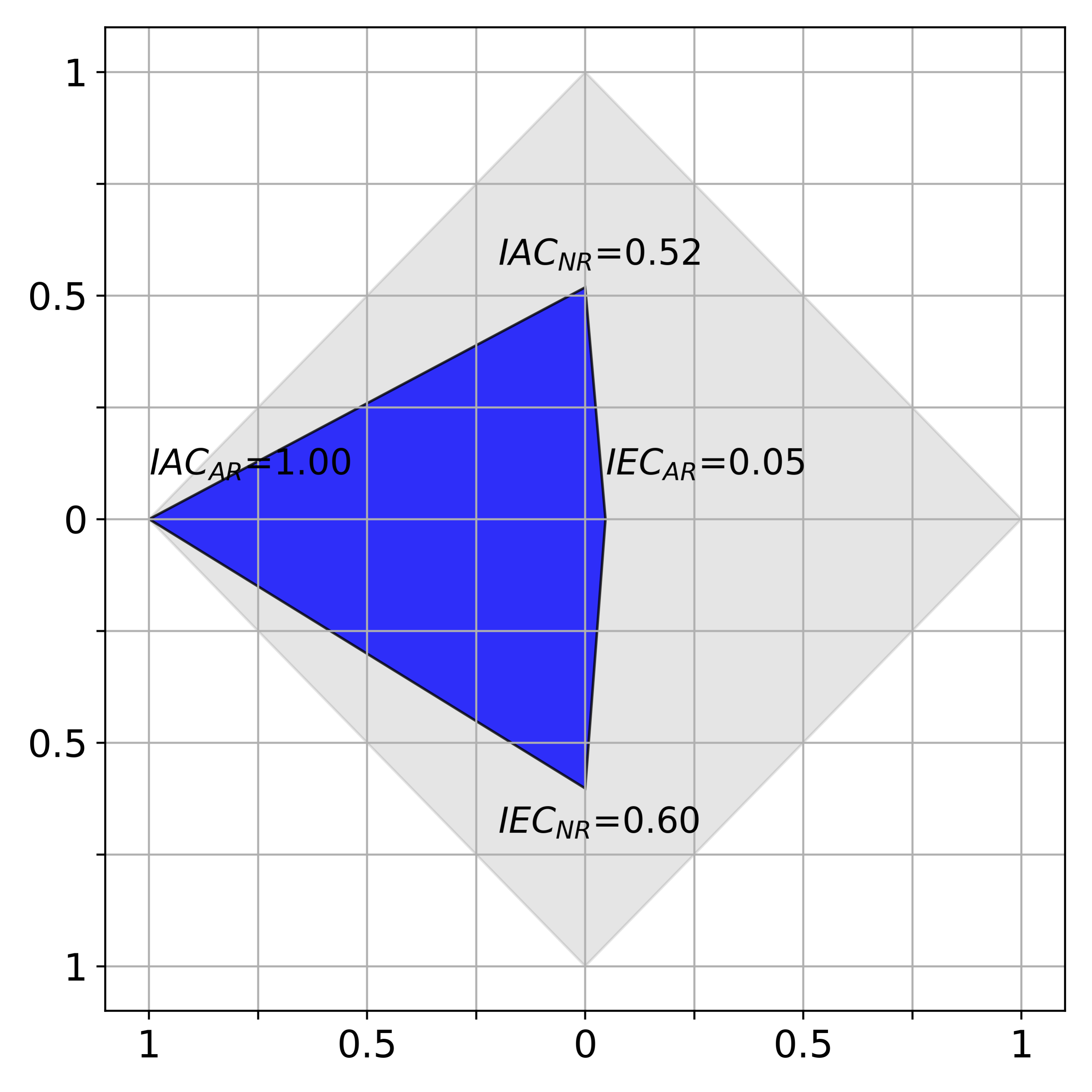}
        \caption{}
        \label{fig:irof}
    \end{subfigure}
    \caption{Results of the evaluation of the evaluation protocol on the \as{ben} dataset \cite{clasen2024reben} for the faithfulness metrics: a) \textit{\al{fe}}; and b) \textit{\al{irof}}.}
    \label{fig:faithfulness_single_area}
\end{figure}
As shown in Fig.~\ref{fig:mc_each_dataset_fig1}, the mean MC scores for the faithfulness category are $0.56$ and $0.60$ for \as{fe} and \as{irof}, respectively. Fig.~\ref{fig:fe} and Fig.~\ref{fig:irof} present the detailed evaluation results for the \as{ben} dataset, which we consider as the perturbation baseline is expected to have a stronger impact for multispectral \as{rs} images.
Similar to the localization metrics, the $\text{IEC}_\text{AR}$ score is notably low with values of 0.28 for \as{fe} and 0.01 for \as{irof}. 
This aligns with methodological findings that faithfulness metrics face similar challenges as localization metrics when an image contains redundant image features for classes that span large parts of the input image. 

In such cases, the prediction certainty is not expected to decrease significantly when important features are removed (see Section~\ref{phantomsec:faithfulness_metrics_problems}). A low $\text{IEC}_\text{AR}$ indicates that the metric scores for the perturbed input are consistently lower than for the unperturbed input. However, when the prediction of the model remains unchanged, faithfulness metrics still assign a high score, keeping the score on the unperturbed input high and consequently leading to low $\text{IEC}_\text{AR}$ values.
On the contrary, the $\text{IAC}_\text{AR}$ scores for \as{fe} and \as{irof} are high, with values of $0.82$ and $1.0$, respectively. This suggests that the distribution of scores remains unchanged even under disruptive perturbations. One possible explanation is a distribution shift caused by \as{ood} samples introduced through the perturbation, which prevents the faithfulness scores from effectively capturing the expected changes.

\begin{figure}
    \centering
    \includegraphics[width=0.7\linewidth]{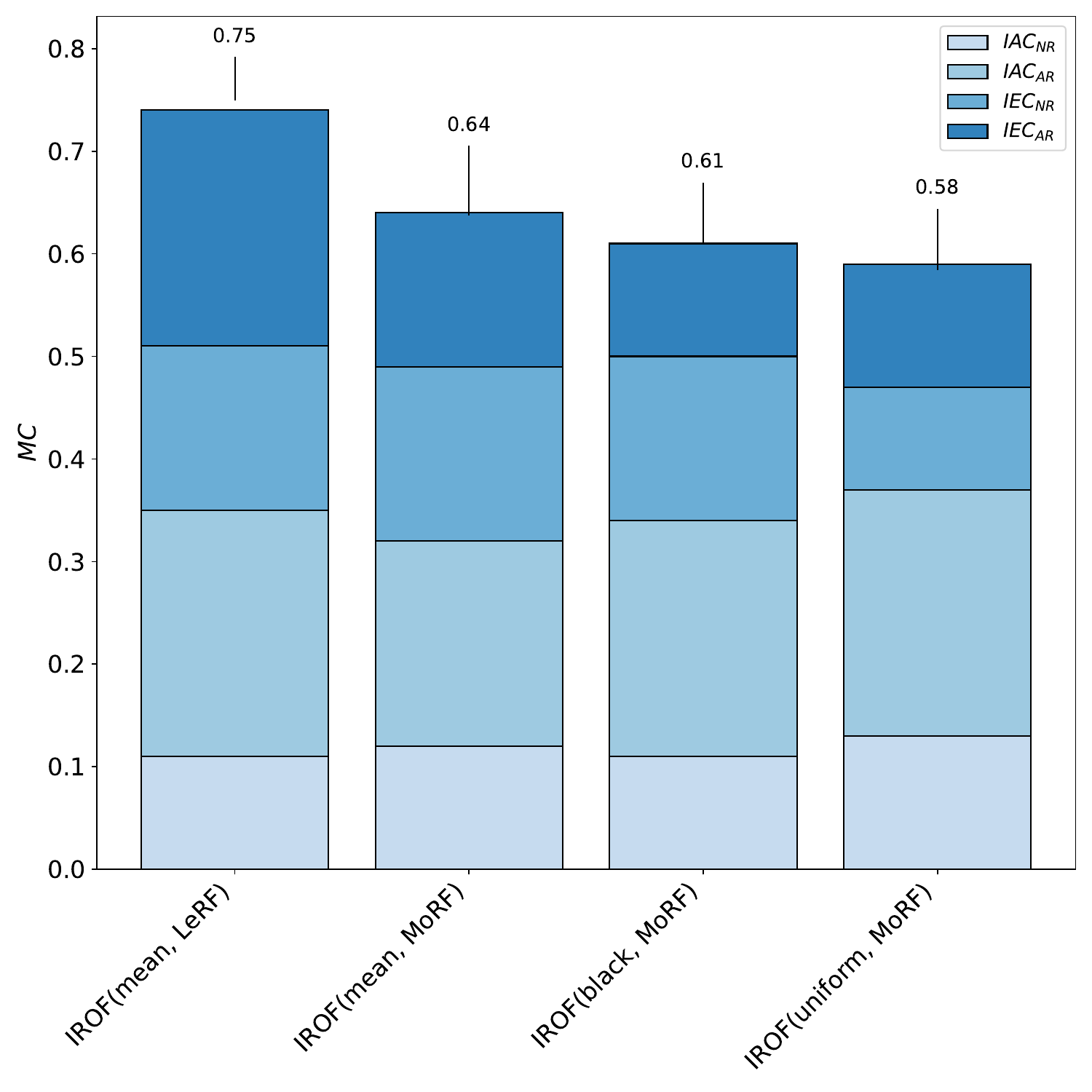}
    \caption{MC Scores of the \acrshort{irof} metric on the \as{ben} dataset \cite{clasen2024reben} under four perturbation configurations: (i) mean-based \acrshort{lerf}, (ii) mean-based \acrshort{morf}, (iii) black-based \acrshort{morf}, and (iv) uniform-based \acrshort{morf}.}

    \label{fig:irof_ablation}
\end{figure}

To further investigate how perturbation baselines and removal strategies influence the reliability of faithfulness metrics on multispectral images, we conduct an ablation study using the \acrshort{irof} metric with four configurations: i) mean-based \acrshort{lerf}; ii) mean-based \acrshort{morf}; iii) black-based \acrshort{morf}; and iv) uniform-based \acrshort{morf} (Fig.\ref{fig:irof_ablation}). The results show that combining the mean baseline with \acrshort{lerf} yields the highest reliability (0.75), followed by mean-based \acrshort{morf} (0.64), black-based \acrshort{morf} (0.61), and uniform-based \acrshort{morf} (0.58). As discussed in Section~\ref{phantomsec:faithfulness_metrics_problems}, removing the most relevant pixels first can fail to capture meaningful changes in prediction certainty, particularly when large image regions contain similar predictive information. In contrast, the \acrshort{lerf} approach preserves critical features until later perturbation steps, leading to a more balanced evaluation. Moreover, employing a mean-value baseline reduces the likelihood of generating \as{ood} samples, further enhancing the reliability of the faithfulness scores.
\subsection{Evaluation of the considered feature attribution methods}
\begin{table*}[ht]
    \centering
    \caption{Average metric scores obtained by the considered explanation methods for BEN-v2.0 \cite{clasen2024reben}.}

    \begin{tabular}{l|cc|cc|cc|cc|cc}
    \toprule
    & 
    \multicolumn{2}{c}{Faithfulness $\uparrow$} & 
    \multicolumn{2}{c}{Robustness $\uparrow$} & 
    \multicolumn{2}{c}{Localization $\uparrow$} &  
    \multicolumn{2}{c}{Complexity $\uparrow$} & 
    \multicolumn{2}{c}{Randomization $\uparrow$} \\
     Explanation Method & 
    \as{fe} \cite{alvarez_melis_towards_2018} & 
    \as{irof} \cite{rieger_irof_2020} & 
    \as{as} \cite{yeh_fidelity_2019} & 
    \as{lle} \cite{alvarez_melis_towards_2018} &  
    \as{tki} \cite{theiner_interpretable_2022} & 
    \as{rra} \cite{arras2022clevr} &  
    \as{sp} \cite{chalasani_concise_2020} & 
    \as{co} \cite{bhatt_evaluating_2020} &  
    \as{rl} \cite{sixt_when_2020} & 
    \as{mprt} \cite{hedstrom2024sanity} \\
    \midrule
    \as{deeplift} \cite{shrikumar_learning_2017}  & 0.66 & \textbf{0.59} & 0.49 & \textbf{1.00} & 0.49 & \textbf{0.48} & \textbf{0.60} & 0.30 & 0.40 & 0.49 \\
    \as{gradcam}  \cite{selvaraju2017grad}   & \textbf{0.73} & 0.56 & \textbf{0.88} & 0.99 & \textbf{0.50} & 0.43 & 0.46 & 0.47 & 0.81 & \textbf{0.67} \\
    \as{lime} \cite{ribeiro_why_2016}      & 0.61 & 0.54 & 0.50 & 0.05 & 0.49 & 0.44 & 0.58 & 0.46 & 0.66 & 0.26 \\
    \as{lrp} \cite{bach2015pixel}       & 0.61 & 0.51 & 0.71 & 0.99 & 0.48 & 0.45 & 0.55 & 0.20 & 0.23 & 0.25 \\
    \as{occlusion} \cite{zeiler_visualizing_2013}  & 0.64 & 0.48 & 0.87 & 0.71 & 0.49 & 0.44 & 0.35 & \textbf{0.52} & 0.74 & 0.42 \\
    Random baseline & 0.39 & 0.44 & 0.85 & 0.60 & 0.47 & 0.42 & 0.07 & 0.00 & \textbf{0.96} & 0.26 \\

    \bottomrule
    \end{tabular}

    \label{tab:average_metric_scores_ben}
\end{table*}
\begin{table*}[ht]
    \centering
    \caption{Average metric scores obtained by the considered explanation methods for DG-ML \cite{DeepGlobe18}.}

    \begin{tabular}{l|cc|cc|cc|cc|cc}
    \toprule
    & 
    \multicolumn{2}{c}{Faithfulness $\uparrow$} & 
    \multicolumn{2}{c}{Robustness $\uparrow$} & 
    \multicolumn{2}{c}{Localization $\uparrow$} &  
    \multicolumn{2}{c}{Complexity $\uparrow$} & 
    \multicolumn{2}{c}{Randomization $\uparrow$} \\
     Explanation Method & 
    \as{fe} \cite{alvarez_melis_towards_2018} & 
    \as{irof} \cite{rieger_irof_2020} & 
    \as{as} \cite{yeh_fidelity_2019} & 
    \as{lle} \cite{alvarez_melis_towards_2018} &  
    \as{tki} \cite{theiner_interpretable_2022} & 
    \as{rra} \cite{arras2022clevr} &  
    \as{sp} \cite{chalasani_concise_2020} & 
    \as{co} \cite{bhatt_evaluating_2020} &  
    \as{rl} \cite{sixt_when_2020} & 
    \as{mprt} \cite{hedstrom2024sanity} \\
    \midrule
    \as{deeplift} \cite{shrikumar_learning_2017}  & 0.67 & 0.65 & 0.41 & \textbf{1.00} & 0.62 & 0.45 & \textbf{0.71} & 0.48 & 0.46 & \textbf{0.66} \\
    \as{gradcam}  \cite{selvaraju2017grad}   & 0.65 & 0.65 & 0.68 & 0.96 & \textbf{0.74} & \textbf{0.48} & 0.43 & 0.42 & 0.83 & 0.49 \\
    \as{lime} \cite{ribeiro_why_2016}   & 0.53 & 0.66 & 0.66 & 0.56 & 0.64 & 0.44 & 0.52 & \textbf{0.50} & 0.72 & 0.23 \\
    \as{lrp} \cite{bach2015pixel}       & 0.69 & 0.66 & 0.61 & 0.99 & 0.64 & 0.47 & 0.54 & 0.26 & 0.22 & 0.42 \\
    \as{occlusion} \cite{zeiler_visualizing_2013}  & \textbf{0.75} & \textbf{0.68} & 0.69 & 0.43 & 0.69 & 0.47 & 0.31 & 0.40 & 0.82 & 0.34 \\
    Random baseline & 0.44 & 0.57 & \textbf{0.84} & 0.78 & 0.59 & 0.37 & 0.19 & 0.03 & \textbf{0.97} & 0.36 \\

    \bottomrule
    \end{tabular}

    \label{tab:average_metric_scores_dg}
\end{table*}
\begin{table*}[ht]
    \centering
    \caption{Average metric scores obtained by the considered explanation methods for FBP-ML \cite{FBP2023}.}
    \label{tab:average_metric_scores_fbp}
    
    \begin{tabular}{l|cc|cc|cc|cc|cc}
    \toprule
    & 
    \multicolumn{2}{c}{Faithfulness $\uparrow$} & 
    \multicolumn{2}{c}{Robustness $\uparrow$} & 
    \multicolumn{2}{c}{Localization $\uparrow$} &  
    \multicolumn{2}{c}{Complexity $\uparrow$} & 
    \multicolumn{2}{c}{Randomization $\uparrow$} \\
     Explanation Method & 
    \as{fe} \cite{alvarez_melis_towards_2018} & 
    \as{irof} \cite{rieger_irof_2020} & 
    \as{as} \cite{yeh_fidelity_2019} & 
    \as{lle} \cite{alvarez_melis_towards_2018} &  
    \as{tki} \cite{theiner_interpretable_2022} & 
    \as{rra} \cite{arras2022clevr} &  
    \as{sp} \cite{chalasani_concise_2020} & 
    \as{co} \cite{bhatt_evaluating_2020} &  
    \as{rl} \cite{sixt_when_2020} & 
    \as{mprt} \cite{hedstrom2024sanity} \\
    \midrule

    \as{deeplift} \cite{shrikumar_learning_2017}   & 0.75 & 0.59 & 0.45 & \textbf{1.00} & 0.46 & 0.49 & \textbf{0.70} & 0.43 & 0.29 & \textbf{0.69} \\
    \as{gradcam}  \cite{selvaraju2017grad}   & 0.75 & \textbf{0.62} & 0.75 & 0.99 & \textbf{0.48} & \textbf{0.50} & 0.57 & \textbf{0.85} & 0.71 & 0.46 \\
    \as{lime} \cite{ribeiro_why_2016}      & 0.65 & 0.55 & 0.56 & 0.41 & 0.45 & 0.45 & 0.55 & 0.51 & 0.67 & 0.16 \\
    \as{lrp} \cite{bach2015pixel}       & 0.69 & 0.58 & 0.53 & 0.99 & 0.43 & 0.47 & 0.65 & 0.28 & 0.29 & 0.26 \\
    \as{occlusion} \cite{zeiler_visualizing_2013}  & \textbf{0.81} & 0.58 & 0.69 & 0.81 & 0.46 & 0.48 & 0.50 & 0.48 & 0.73 & 0.25 \\
    Random baseline & 0.37 & 0.39 & \textbf{0.85} & 0.45 & 0.45 & 0.41 & 0.06 & 0.00 & \textbf{0.98} & 0.17 \\

    \bottomrule
    \end{tabular}
\end{table*}
\noindent
In this subsection, we assess the considered feature attribution methods for \as{rs} image scene classification using the considered explanation metrics grouped by category. Tables \ref{tab:average_metric_scores_ben}, \ref{tab:average_metric_scores_dg}, and \ref{tab:average_metric_scores_fbp} report the average metric scores from each dataset.
Faithfulness is assessed using the \as{fe} and \as{irof} metrics. As observed in the tables, \as{occlusion} achieves the highest \as{fe} score for the \as{dg} and \as{fbp} datasets, whereas \as{gradcam} performs best for the \as{ben} dataset. For the \as{irof} metric, \as{deeplift}, \as{occlusion}, and \as{gradcam} attain the highest scores for \as{ben}, \as{dg} and \as{fbp}, respectively. Averaged across all datasets, \as{occlusion} outperforms all other methods in the \as{fe} metric with an average score of 0.73, while \as{gradcam} and \as{deeplift} achieve the highest \as{irof} score of 0.61. Both metrics exhibit low variance across datasets.
Robustness is measured using the \as{as} and \as{lle} metrics. In the \as{as} metric, the random baseline outperforms all methods on the \as{dg} and \as{fbp} datasets. This is because the \as{as} metric assesses how much the explanation changes on average when noise is introduced. Given a sufficiently large sample size, a uniformly drawn distribution remains relatively stable under such perturbations, leading to higher scores for the random baseline. Since the \as{lle} metric measures the maximum change, the random baseline performs worse. Among explanation methods, \as{gradcam} achieves the highest \as{as} score for \as{ben} and \as{fbp}, while \as{occlusion} performs best for \as{dg}. For the \as{lle} metric, \as{deeplift} achieves the highest score across all datasets, closely followed by \as{gradcam} and \as{lrp}. Both perturbation-based methods (\as{lime} and \as{occlusion}) perform worse in the \as{lle} metric.

Localization is evaluated using the \as{tki} and \as{rra} metrics. Across all datasets and metrics, except for \as{rra} on \as{ben}, \as{gradcam} performs best. The variance across the scores for the different datasets and methods remains low, except for the \as{tki} metric in the \as{dg} dataset, which exhibits higher variability.
Complexity is assessed using the \as{sp} and \as{co} metrics. For the \as{sp} metric, \as{deeplift} achieves the highest performance across all datasets. The results for the \as{co} metric vary across datasets; averaged over all datasets, \as{gradcam} yields the best performance with an average score of 0.58. The random baseline performs poorly, as expected, given that the \as{co} metric is defined as the entropy of the explanation, which reaches its maximum when the explanation is uniformly distributed.
Randomization is quantified using the \as{rl} and \as{mprt} metrics. The random baseline achieves the highest \as{rl} score across all datasets, with an average performance of 0.97. This aligns with the findings of \cite{binder2023shortcomings}, which demonstrate that randomized attribution maps can achieve high scores in top-down randomization metrics. 
While this outcome exposes a limitation of the RL metric, it remains valuable in identifying explanation methods that are largely independent of the model and therefore unreliable.
For the \as{mprt} metric, \as{gradcam} achieves the best result on the \as{ben} dataset, while \as{deeplift} performs best on \as{dg} and \as{fbp}.

\begin{figure}[h]
    \centering
    \includegraphics[width=1\linewidth]{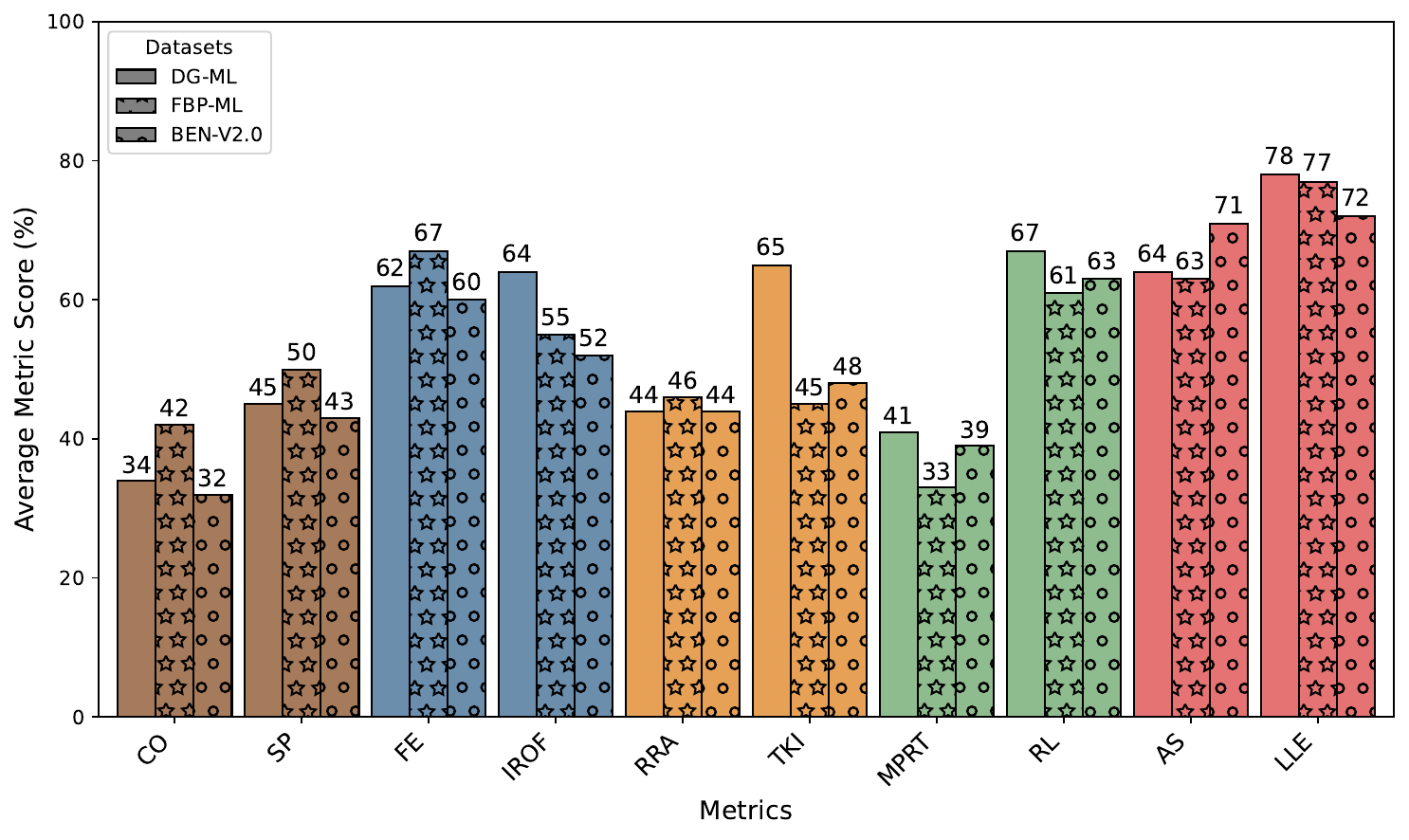}
\caption{Average performance of explanation metrics for each considered dataset, grouped by evaluation category. Each bar corresponds to one metric, with colors indicating the metric category: brown for complexity, blue for faithfulness, orange for localization, green for randomization, and red for robustness.}
    \label{fig:metric-comparison}
\end{figure}

As illustrated in Fig.~\ref{fig:metric-comparison}, the largest dataset-dependent differences occur in the \as{tki} metric, which quantifies how well explanations align with pixel-level reference maps. Specifically, the scores are notably higher for the \as{dg} dataset, likely due to its low number of unique labels and lower average number of labels per image, compared to the other datasets (see Table~\ref{table:ds_characteristics}). 

%

\begin{figure}
    \centering
    \begin{subfigure}{0.49\linewidth}
        \centering
        \includegraphics[width=\linewidth]{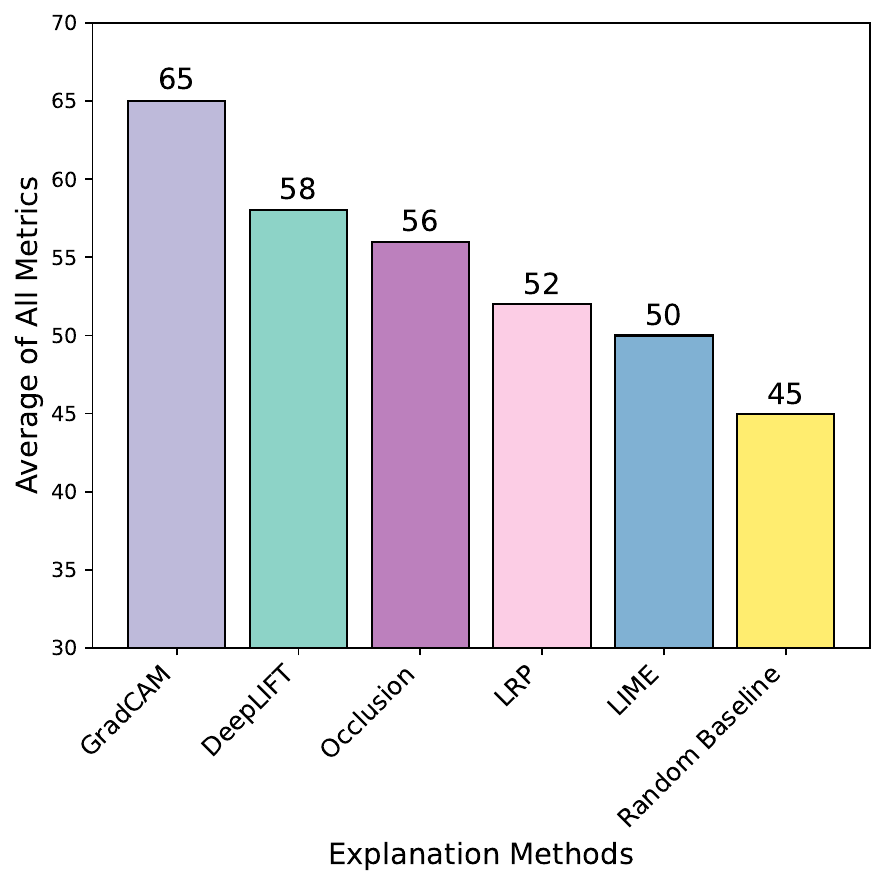}
        \caption{}
        \label{fig:eval_average}
    \end{subfigure}
    \hfill
    \begin{subfigure}{0.49\linewidth}
        \centering
        \includegraphics[width=\linewidth]{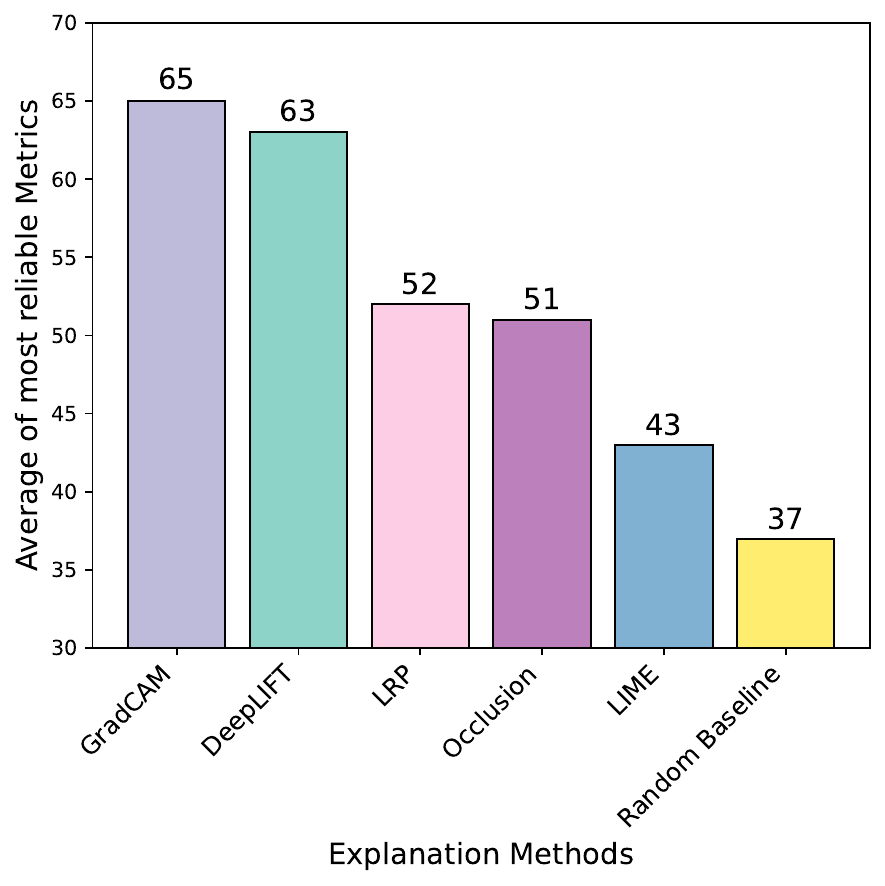}
        \caption{}
        \label{fig:weighted_eval_average}
    \end{subfigure}
    \caption{Comparison of explanation methods based on evaluation metrics. a): normalized average of all the considered metrics across all the considered \as{rs} datasets; b): normalized average of the best-performing metric (\as{co}, \as{irof}, \as{lle}, \as{rra}, \as{mprt}) for each category across all considered \as{rs} datasets.}
    \label{fig:combined_eval}
\end{figure}

Fig.~\ref{fig:eval_average} and Fig.~\ref{fig:weighted_eval_average} show a comparison of the overall and weighted average performances of the evaluated explanation methods across all considered metrics. In both evaluations, \as{gradcam} consistently achieves the highest scores. When weighting by the most reliable metrics, the relative differences between methods become more pronounced. In detail, \as{deeplift} improves its standing, while \as{lime} and the Random Baseline show a substantial decline. This indicates that reliable metrics provide a sharper separation between strong and weak methods, effectively penalizing random explanations and demonstrating the reliability of \as{gradcam}.

Overall, \as{gradcam} emerges as the most versatile explanation method, as it achieves strong performance across multiple evaluation categories. It excels in faithfulness, localization, and randomization, making it particularly well-suited for \as{rs} image scene classification.
\subsection{Evaluation of the computational efficiency}
\label{subsec:efficiency}

\begin{table}[h]
\centering
\caption{Average computation time (in seconds) for metrics and explanation methods on the \hbox{\as{dg} dataset \cite{DeepGlobe18}.}}
\label{tab:combined_time}
\begin{tabular}{llr}
\toprule
\textbf{Group} & \textbf{Name} & \textbf{Time (seconds)} \\
\midrule
\multirow[c]{6}{*}{Explanation Methods} 
& \hbox{\as{deeplift}} \cite{shrikumar_learning_2017} & 0.26 \\
& \hbox{\as{gradcam}} \cite{selvaraju2017grad} & 0.12 \\
& \hbox{\as{lime}} \cite{ribeiro_why_2016} & 28.63 \\
& \hbox{\as{lrp}} \cite{bach2015pixel} & 0.38 \\
& \hbox{\as{occlusion}} \cite{zeiler_visualizing_2013} & 2.04 \\
& Random baseline & 0.04 \\
\midrule
\multirow[c]{10}{*}{Explanation Metrics} 
& \hbox{\as{fe}} \cite{alvarez_melis_towards_2018} & 3.39 \\
& \hbox{\as{irof}} \cite{rieger_irof_2020} & 13.26 \\
& \hbox{\as{as}} \cite{yeh_fidelity_2019} & 12.15 \\
& \hbox{\as{lle}} \cite{alvarez_melis_towards_2018} & 11.97 \\
& \hbox{\as{tki}} \cite{theiner_interpretable_2022} & 0.10 \\
& \hbox{\as{rra}} \cite{arras2022clevr} & 0.16 \\
& \hbox{\as{sp}} \cite{chalasani_concise_2020} & 0.06 \\
& \hbox{\as{co}} \cite{bhatt_evaluating_2020} & 0.06 \\
& \hbox{\as{rl}} \cite{sixt_when_2020} & 0.03 \\
& \hbox{\as{mprt}} \cite{hedstrom2024sanity} & 11.10 \\
\bottomrule
\end{tabular}
\end{table}

\noindent
In this subsection, we analyze the computational efficiency of the considered metrics and explanation methods for \hbox{\as{rs}} scene classification. \hbox{Table~\ref{tab:combined_time}} reports the average runtime required in seconds to compute each explanation method or metric on the \hbox{\as{dg}} dataset.

From the table, one can observe that the average computation time required for the relevancy-propagation-based and gradient-propagation-based methods, such as \hbox{\as{gradcam}}, \hbox{\as{deeplift}}, and \hbox{\as{lrp}}, is below one second. This is because these methods explain the prediction in a single forward pass. In contrast, perturbation-based approaches such as \hbox{\as{lime}} and \hbox{\as{occlusion}} require substantially more computation, as they require multiple forward passes. The runtime of \hbox{\as{lime}} is particularly high (28.63 seconds) due to the need to evaluate perturbed samples and fit a surrogate model. However, it is worth noting that the efficiency of perturbation-based methods depends mostly on their parametrization, as increasing the number of perturbations improves the reliability of the results but proportionally increases computation time.

For explanation metrics, a similar behavior can be observed from the table. Perturbation-based metrics such as \hbox{\as{fe}}, \hbox{\as{irof}}, \hbox{\as{as}}, \hbox{\as{lle}}, and \hbox{\as{mprt}} require several seconds due to repeated recomputation of explanations under perturbations. In contrast, metrics from localization, complexity, and randomization categories, including \hbox{\as{tki}}, \hbox{\as{rra}}, \hbox{\as{sp}}, \hbox{\as{co}}, and \hbox{\as{rl}}, are computed in closed form from a single attribution and thus achieve runtimes below 0.2 seconds. These results highlight that the computational efficiency of explanation metrics is largely dictated by whether perturbation is required.


\section{Conclusion and Discussion}
\label{sec:conclusion_discussion}
\noindent
In this study, for the first time in \as{rs}, we have systematically investigated the suitability of explanation methods and metrics for \as{rs} image scene classification by combining methodological analysis with extensive experiments on three datasets. 
In particular, we analyzed five widely used feature attribution methods: \as{occlusion} \cite{zeiler_visualizing_2013}, \as{lime} \cite{ribeiro_why_2016}, \as{lrp} \cite{bach2015pixel}, \as{deeplift} \cite{shrikumar_learning_2017} and \as{gradcam} \cite{selvaraju2017grad} and ten explanation metrics spanning five categories: faithfulness (\as{fe} \cite{alvarez_melis_towards_2018}, \as{irof} \cite{rieger_irof_2020}), robustness (\as{as} \cite{yeh_fidelity_2019}, \as{lle} \cite{alvarez_melis_towards_2018}), localization (\as{rra} \cite{arras2022clevr}, \as{tki} \cite{theiner_interpretable_2022}), complexity (\as{sp} \cite{chalasani_concise_2020}, \as{co} \cite{bhatt_evaluating_2020}) and randomization-based (\as{mprt} \cite{adebayo_sanity_2018}, \as{rl} \cite{sixt_when_2020}). 

Our methodological analysis shows that perturbation‐based approaches, such as \as{occlusion} and \as{lime}, are critically dependent on the choice of perturbation baseline. In RS scenes, replacing spectral content with a fixed value can generate \as{ood} samples. Furthermore, the redundancy of features across large regions hinders the perturbation function to significantly decrease the prediction of a model, which lowers the reliability of perturbation-based methods. 
Gradient‐propagation methods like \as{gradcam} are limited in their ability to distinguish between multiple instances of the same object in a scene. Relevancy-propagating methods like \as{lrp} tend to underestimate the contribution of LULC classes that span larger areas of a scene.
Regarding evaluation metrics, our findings indicate that faithfulness metrics inherit the limitations of perturbation‐based methods and need to be parametrized accordingly. In contrast, robustness metrics demonstrate greater stability. Localization metrics require accurate pixel-level reference maps. Furthermore, localization and complexity metrics are unreliable when a single class spans a large part of the scene. Randomization metrics are often independent of the data.
To support our methodological findings, we conducted two sets of experiments on three \as{rs} datasets: \as{ben} \cite{clasen2024reben}, \as{dg} \cite{DeepGlobe18}, and \as{fbp} \cite{FBP2023}. In the first set of experiments, we meta-evaluated ten explanation metrics to assess their reliability. We observe that metrics from the robustness and randomization categories are reliable. As expected from the methodological analysis, metrics from the faithfulness and localization categories are less reliable.
In the second set of experiments, we use the most reliable explanation metrics to determine which explanation methods are best suited for \as{rs} image scene classification. Among the evaluated methods, \as{gradcam} emerges as the most robust feature attribution method, demonstrating relatively stable performance across different datasets and evaluation metrics. However, it is important to note that no single method has performed optimally across all evaluation criteria.

Following our methodological and experimental analysis, we develop guidelines for selecting suitable explanation metrics and explanation methods for scene classification in \as{rs}: 

\begin{enumerate}
    \item We recommend using robustness and randomization metrics due to their reliability across datasets and explanation methods.

    \item Faithfulness metrics should be used cautiously, as they depend on perturbation baselines and may not reliably reflect model behavior. To mitigate this issue, we recommend using the \as{lerf} removal strategy, which helps to maintain reliability by gradually removing less relevant features first. Furthermore, we recommend utilizing the mean value as a perturbation baseline.

    \item We suggest using localization metrics only when accurate pixel-level reference maps are available. In particular, we recommend using more advanced localization metrics that also account for the size of the target class. Complexity metrics should be interpreted carefully, as they depend strongly on the spatial extent of the target class. Similar to localization metrics, complexity metrics could also be modified to account for the target class size.

    \item We recommend using \as{gradcam} as a general baseline due to its versatility, speed, and ease of interpretation. Among the evaluated methods, it consistently achieves high scores across multiple metric categories.

\end{enumerate}

Beyond methodological contributions, our findings have direct implications for operational applications. For instance, reliable explanation methods are crucial in maritime monitoring, where RS scene classification is used for maritime surveillance and decision support. Moreover, reliable attribution maps can serve as pseudo labels in tasks such as tree species classification \mbox{\cite{ahlswede2022weakly}}.

As a final remark, we emphasize that the evaluation protocol can not measure the intrinsic validity of explanation metrics, since no ground truth for explanations exists \mbox{\cite{hedstrom2023meta}}. It can probe the reliability and comparative behavior of metrics, yet reliability alone does not imply validity \mbox{\cite{binder2023shortcomings}}. Evaluation outcomes should therefore be regarded as indicative rather than definitive, and the theoretical foundations of the metrics also have to be taken into account.

As a future work, we plan to refine explanation metrics specifically designed for \as{rs} image scene classification. In particular, we want to develop perturbation strategies that better accommodate spectral variability and feature redundancy, which may enhance the reliability of faithfulness metrics and, by extension, perturbation-based explanation methods. Although our study concentrated on feature attribution methods commonly used in \as{rs} (while being originally designed for natural images), a promising direction for future research is to develop explanation methods and metrics tailored specifically to the characteristics of \as{rs} images. Another important direction is to evaluate more recent explanation approaches, such as concept-based XAI, and to extend the evaluation framework accordingly.


%


\ifCLASSOPTIONcaptionsoff
  \newpage
\fi



\bibliographystyle{IEEEtran}
\bibliography{bibtex/bib/IEEEabrv, bibtex/bib/main}
%



%

\newpage
\begin{IEEEbiography}[{\includegraphics[width=1in,height=1.25in,clip,keepaspectratio]{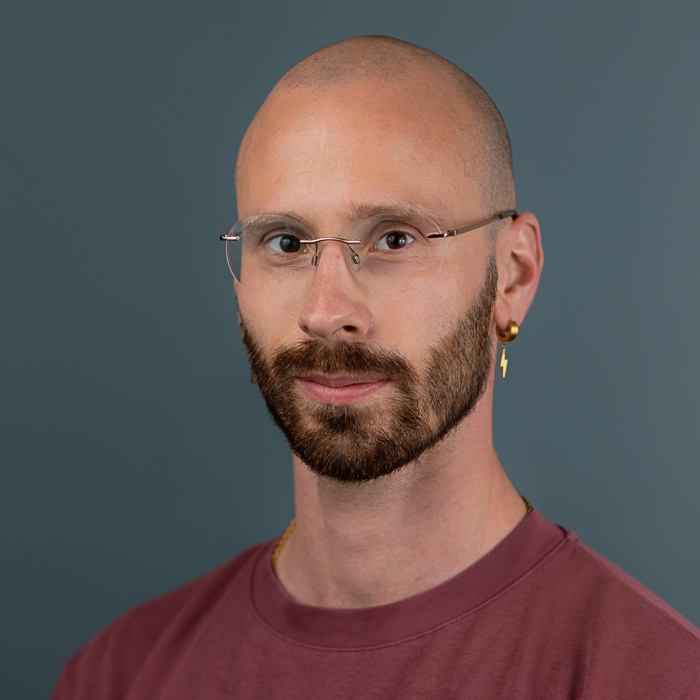}}]{Jonas Klotz}
(Member, IEEE) 
received the M.Sc. degree in computer science from Technische Universität (TU) Berlin, Berlin, Germany, in 2024. He is currently pursuing a Ph.D. in machine learning at the Berlin Institute for the Foundations of Learning and Data (BIFOLD), with the Remote Sensing and Image Analysis (RSiM) Group. His research interests center on the intersection of explainable AI and remote sensing.
\end{IEEEbiography}
%
\begin{IEEEbiography}[{\includegraphics[width=1in,height=1.25in,clip,keepaspectratio]{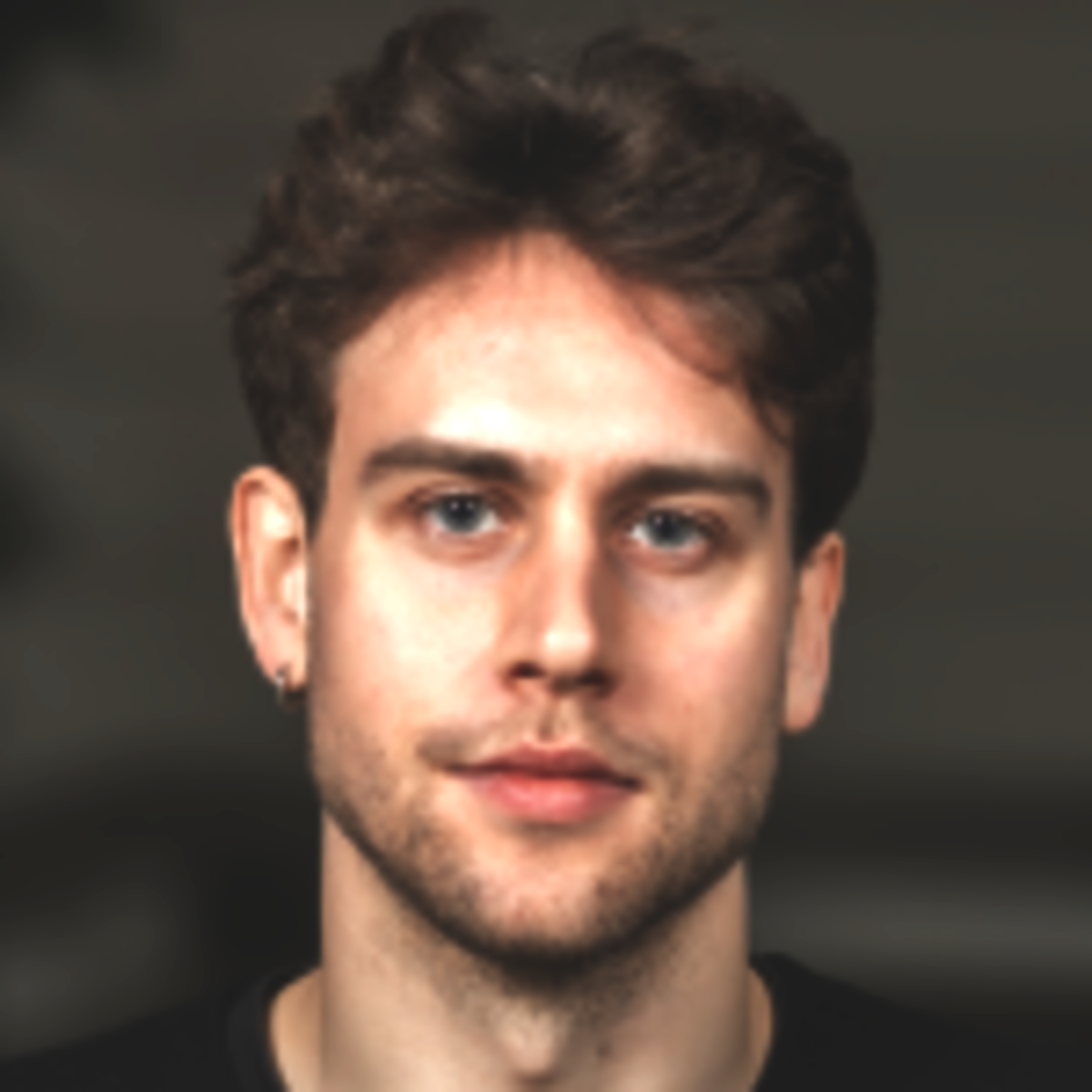}}]{Tom Burgert}
(Member, IEEE) 
received the M.Sc. degree in computer science from Technische Universität (TU) Berlin, Berlin, Germany, in 2022. He is currently pursuing a Ph.D. in machine learning with the Remote Sensing and Image Analysis (RSiM) group at the Berlin Institute for the Foundations of Learning and Data (BIFOLD) and TU Berlin. His research interests evolve around self-supervised learning and explaining deep neural networks for computer vision and remote sensing.
\end{IEEEbiography}
%
\begin{IEEEbiography}[{\includegraphics[width=1in,height=1.25in,clip,keepaspectratio]{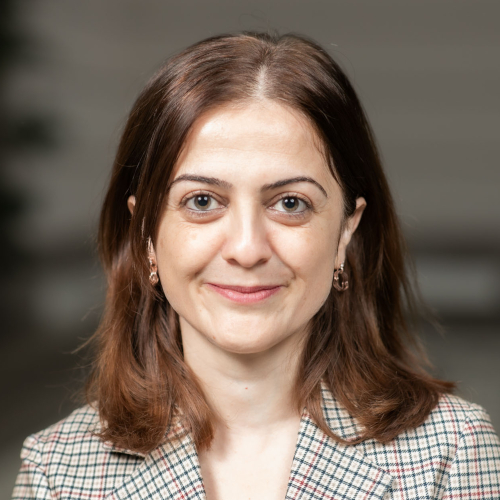}}]{Begüm Demir} (Senior Member, IEEE) (demir@tu-berlin.de) (S'06-M'11-SM'16) received the B.Sc., M.Sc., and Ph.D. degrees in electronic and telecommunication engineering from Kocaeli University, Kocaeli, Turkey, in 2005, 2007, and 2010, respectively.
She is currently a Full Professor and the founder head of the Remote Sensing Image Analysis (RSiM) group at the Faculty of Electrical Engineering and Computer Science, TU Berlin and the head of the Big Data Analytics for Earth Observation research group at the Berlin Institute for the Foundations of Learning and Data (BIFOLD). Her research activities lie at the intersection of machine learning, remote sensing and signal processing. Specifically, she performs research in the field of processing and analysis of large-scale Earth observation data acquired by airborne and satellite-borne systems. She was awarded by the prestigious ‘2018 Early Career Award’ by the IEEE Geoscience and Remote Sensing Society for her research contributions in machine learning for information retrieval in remote sensing. In 2018, she received a Starting Grant from the European Research Council (ERC) for her project “BigEarth: Accurate and Scalable Processing of Big Data in Earth Observation”. She is an IEEE Senior Member and Fellow of European Lab for Learning and Intelligent Systems (ELLIS).
Prof. Demir is a Scientific Committee member of several international conferences and workshops. She is a referee for several journals such as the PROCEEDINGS OF THE IEEE, the IEEE TRANSACTIONS ON GEOSCIENCE AND REMOTE SENSING, the IEEE GEOSCIENCE AND REMOTE SENSING LETTERS, the IEEE TRANSACTIONS ON IMAGE PROCESSING, Pattern Recognition, the IEEE TRANSACTIONS ON CIRCUITS AND SYSTEMS FOR VIDEO TECHNOLOGY, the IEEE JOURNAL OF SELECTED TOPICS IN SIGNAL PROCESSING, the International Journal of Remote Sensing), and several international conferences. Currently she is an Associate Editor for the IEEE GEOSCIENCE AND REMOTE SENSING MAGAZINE.
\end{IEEEbiography}




\end{document}